\documentclass[lettersize,journal]{IEEEtran}
\usepackage{threeparttable }
\usepackage{amsmath,amsfonts}
\usepackage{algorithmic}
\usepackage{algorithm}
\usepackage{array}
\usepackage[caption=false,font=normalsize,labelfont=sf,textfont=sf]{subfig}
\usepackage{textcomp}
\usepackage{stfloats}
\usepackage{url}
\usepackage{verbatim}
\usepackage{graphicx}
\usepackage{cite}
\usepackage{amsmath}

\usepackage{booktabs}
\usepackage{multirow}
\usepackage{tabularx}
\usepackage{hyperref}
\hypersetup{hypertex=true,
colorlinks=true,
linkcolor=blue,
anchorcolor=blue,
citecolor=blue}
\begin{document}

\title{ErasableMask: A Robust and Erasable Privacy Protection Scheme against Black-box Face Recognition Models}

\author{Sipeng Shen, Yunming Zhang, Dengpan Ye,~\IEEEmembership{Member,~IEEE}, Xiuwen Shi, Long Tang, Haoran Duan, Yueyun Shang, Zhihong Tian,~\IEEEmembership{Senior Member,~IEEE}
\thanks{Sipeng Shen, Yunming Zhang, Dengpan Ye, Xiuwen Shi, Long Tang, and Haoran Duan are with the Key Laboratory of Aerospace Information Security and Trusted Computing, Ministry of Education, School of Cyber Science and Engineering, Wuhan University. E-mail: {\{sipeng, zhangyunming, yedp, shixiuwen, l$\_$tang\, haoraod}@whu.edu.cn. Yueyun Shang is with School of Mathematics and Statistics, South-Central University for Nationalities, Wuhan 430070, P.R. China. E-mail: 36214001@qq.com. Zhihong Tian is with Cyberspace Institute of Advanced Technology, Guangzhou University,  Guangdong Key Laboratory of Industrial Control System Security, and  Huangpu Research School of Guangzhou University. E-mail: tianzhihong@gzhu.edu.cn.
Sipeng Shen and Yunming Zhang contributed equally to this work.\\
(Corresponding author: Dengpan Ye, Yueyun Shang.)}}
\markboth{Journal of \LaTeX\ Class Files,~Vol.~14, No.~8, August~2021}%
{Shell \MakeLowercase{\textit{et al.}}: A Sample Article Using IEEEtran.cls for IEEE Journals}

\maketitle
\begin{abstract}
                While face recognition (FR) models have brought remarkable convenience in face verification and identification, they also pose substantial privacy risks to the public. Existing facial privacy protection schemes usually adopt adversarial examples to disrupt face verification of FR models.
                However, these schemes often suffer from weak transferability against black-box FR models and permanently damage the identifiable information that cannot fulfill the requirements of authorized operations such as forensics and authentication.
                To address these limitations, we propose \textbf{ErasableMask}, a robust and erasable privacy protection scheme against black-box FR models. Specifically, via rethinking the inherent relationship between surrogate FR models, ErasableMask introduces a novel meta-auxiliary attack, which boosts black-box transferability by learning more general features in a stable and balancing optimization strategy. It also offers a perturbation erasion mechanism that supports the erasion of semantic perturbations in protected face without degrading image quality. To further improve performance, ErasableMask employs a curriculum learning strategy to mitigate optimization conflicts between adversarial attack and perturbation erasion.  Extensive experiments on the CelebA-HQ and FFHQ datasets demonstrate that ErasableMask achieves the state-of-the-art performance 
                in transferability, achieving over \textbf{72\%} mean confidence in commercial FR systems. Moreover, ErasableMask also exhibits outstanding perturbation erasion performance, achieving over \textbf{90\%} erasion success rate.
\end{abstract}


\section{Introduction}
        \IEEEPARstart{W}{ith} the advancement of deep learning, face recognition (FR) technology has become deeply integrated into daily life~\cite{ding2015robust,zhong2021dynamic,chen2015face,choi2010collaborative}. Users frequently share personal facial images on social media platforms such as Twitter, Facebook, and LinkedIn, which provides great convenience for trusted authorities (TAs) in performing face verification. However, this convenience comes with significant privacy risks. Attackers can match these images with existing biometric databases to accurately identify individuals, facilitating serious criminal activities~\cite{larson2018pixel}. As a result, there is an urgent need for a robust scheme that can address the misuse and threats posed by FR technology while effectively preserving the usability of the information for TAs.
          \begin{figure}
            \centering
            \includegraphics[width=0.5\textwidth]{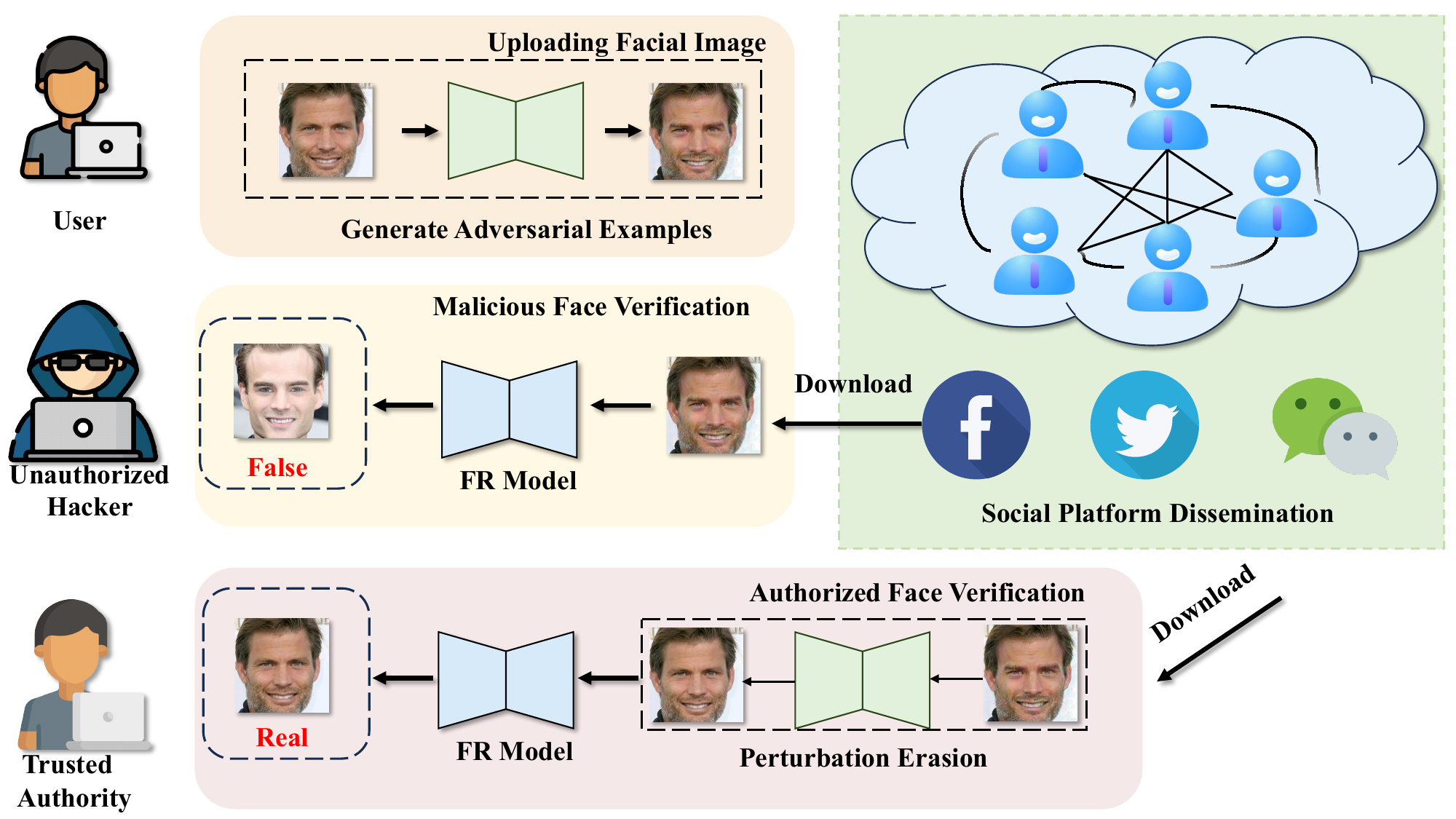}
            \caption{ErasableMask protection scenario: Users generate protected faces and then publicly share them on social platforms. Unauthorized hackers fail to conduct face verification. However, trusted authority can utilize ErasableMask's erasion module to obtain clean examples that are nearly identical to the original face, allowing to conduct face verification accurately.}
        \label{Fig1}
        \end{figure} 
        
        Existing adversarial examples~\cite{advfaces,komkov2021advhat,dong2019efficient,sharif2016accessorize,yang2021attacks,semanticadv} against FR models can serve as a feasible solution. However, they suffer from the following challenges:   
        1). 
        \textbf{Existing schemes exhibit weak transferability towards strictly black-box FR models. }
        Some transferable adversarial example schemes have been proposed in recent year~\cite{jia2022adv, hu2022protecting,shamshad2023clip2protect,liu2024adv,hu2024towards}, which simply employ ensemble attack for all surrogate models. Unfortunately, this often leads to overfitting of the surrogate white-box models\cite{yin2021adv}, resulting in weak transferability. 
        In ensemble attacks, models often prioritize fitting the more easily attackable surrogate models before gradually adapting to the relatively harder-to-attack ones. This observation indicates an imbalance in the optimization process. Consequently, even if the ensemble attack loss converges, the model tends to overfit to the surrogate models and fails to learn robust and general features.
         This raises the question: \textit{Can the model learn more robust and general features if it can learn in a stable and balancing attack strategy?}
        2). 
        \textbf{The adversarial performance permanently disrupts face verification of trusted authority for forensics and authentication. }
        The most obvious obstacle encountered by existing schemes is their permanently damage to the identifiable
        information present in the source faces. This severely hinders the ability of TAs to accurately perform face verification. Reversible Adversarial Examples (RAEs) can serve as one of potential strategies for TA's demands~\cite{xiong2023black,liu2023unauthorized,zhang2022self,xie2024reversible}. 
        Most of them adopt a reversible data hiding strategy to fulfil recovery of original images~\cite{yin2019reversible,liu2023unauthorized,xiong2023black}. 
        However, RAEs exhibit several limitations. They fail to address the robustness of recovery that cannot tackle with various image process methods in real-world scenarios and overlook facial identity protection, leaving a gap in facial privacy protection scenarios. Additionally, existing RAEs typically combine pixel-level adversarial perturbations with separate data hiding techniques. This raises the question: \textit{Would it be possible to design semantic perturbations in facial privacy protection scenarios that are inherently self-erasable?}
    
        To resolve the questions mentioned above, we present a novel facial privacy protection scheme called ErasableMask, which utilizes facial attributes as conditions to generate transferable and erasable adversarial examples with strong robustness to tackle with various image process methods. Specifically, ErasableMask solves the above question through the following strategies: 
        
        \textit{To tackle with question 1.} Existing studies have demonstrated that auxiliary learning can serve as a promising balancing strategy, which enhances generalization of the primary task by learning additional relevant features obtained in the sharing of features with auxiliary tasks~\cite{liu2019self,chen2024joint}. This motivates us to think about whether the model can learn a more transferable attack through this stable and balancing strategy, where attacking one specific surrogate FR model serves as a primary task and fine-grained fine-tuned the model with feedback from other surrogate FR models as auxiliary tasks. But there exists a question of how to transfer the strategy learned from the primary task to the auxiliary tasks, while also allowing the auxiliary tasks to provide beneficial feedback.  
        Recently, meta-learning has been utilized to improve the optimization compatibility between different tasks~\cite{finn2017model,shao2020regularized}. 
        As a result, we introduce a novel meta-auxiliary attack to learn more robust and general features that boost transferability. Specifically, we split surrogate FR models into a primary task and several auxiliary tasks before each attack, then use the performance of auxiliary tasks in meta-test to further fine-grained fine-tuned optimization direction of the primary task in meta-train, which is a form of meta-learning with double gradient. 
         
         \textit{To tackle with question 2.} 
         One direct method to enable self-erasable capability is deploying a restorer, allowing it to be tightly coupled with the generator and trained end-to-end with it. 
         However, there exists a problem that the clean-domain information in protected faces is severely damaged, resulting in a difficulty for restorer to obtain enough information that facilitate perturbation erasion. This inspires us to consider whether we can retain enough clean-domain information that can guide perturbation erasion. As a result, we employ a clean-domain information injection strategy which embeds additional source face information via a copy of generator to improve perturbation erasion performance. Moreover, the adversarial generation and erasion of semantic perturbations are a set of conflicting optimization objectives. And to tackle with real-world transmission standards, ErasableMask also needs to hold robustness both in adversarial attack and perturbation erasion, which can be fulfilled by a differential noise pool~\cite{jia2021mbrs}. Directly deploying this noise pool will undoubtedly increase the difficulty of training for this conflicting task. This prompts us to think about making the ErasableMask generator and restorer acquire adversarial, erasable, and robust capabilities step by step to mitigate conflicts. Thus, we adopt a three-stage curriculum learning\cite{curriculum} strategy. Specifically, we allow ErasableMask to solely focus on attribute modification strategy in stage 1 and deploy the noise pool in the last two phases (introducing robust adversarial example in stage 2 and robust perturbation erasion in stage 3).

        To the best of our knowledge, ErasableMask is the first self-erasable facial privacy protection scheme. 
        As shown in Fig.~\ref{Fig1}, ErasableMask strikes a balance between facial privacy protection and information usability. Compared with existing works in Table.~\ref{compare_research}, ErasableMask supports self-erasable ability and strong robustness both in attack and recovery.
        Overall, ErasableMask makes the following contributions:
    \begin{enumerate}
         \item We design a meta-auxiliary attack strategy that balances attack stability and generalization. This strategy encourages the model to learn transferable features, significantly improving the attack performance on black-box face recognition (FR) systems.

        \item We introduce a curriculum learning-based optimization scheme that progressively aligns adversarial attack objectives with perturbation erasion goals, mitigating the inherent conflicts between them and leading to step-wise improvement in both attack and erasion capability.

        \item Extensive experiments demonstrate that ErasableMask achieves state-of-the-art performance across three dimensions: black-box transferability (average confidence over \textbf{72\%} in commercial FR systems), perturbation erasion (\textbf{90\%}+ success rate in black-box scenarios), and robustness (minimal performance drop under common image transformations).

    \end{enumerate}

    \section{Related Work}
	\label{sec.2}
    \begin{table}[!tbph]
        \renewcommand{\arraystretch}{1.0}
        \centering
        \caption{Comparison of works on adversarial examples.}
        \label{compare_research}
        \resizebox{0.5\textwidth}{!}{ 
        \begin{tabular}{c|cccccc}
        \hline
        \multirow{2}{*}{\textbf{Scheme}} &  \multirow{2}{*}{\textbf{Victim}} & \multirow{2}{*}{\textbf{Perturb}} & \multicolumn{2}{c}{\textbf{Recovery}} & \multicolumn{2}{c}{\textbf{Robustness}} \\ \cline{4-7} 
                               &                                                             &                                  & \textbf{Combined} & \textbf{Self-erasable} & \textbf{Attack} & \textbf{Recovery} \\ \hline
        ~\cite{fgsm}                          & Class           & Pixel          &  $\times$     & $\times$               & $\times$             & $\times$                   \\ 
        ~\cite{pgd}                             & Class           & Pixel           & $\times$       & $\times$               & $\times$             & $\times$                   \\ 
        ~\cite{liu2023unauthorized}               & Class           & Pixel           & $\checkmark$      & $\times$               & $\times$             & $\times$                   \\ 
        ~\cite{xiong2023black}                   & Class           & Pixel           & $\checkmark$      & $\times$               & $\checkmark$           & $\times$                   \\ 
        ~\cite{zhang2022self}                  & Class          & Pixel           & $\times$          & $\checkmark$            & $\checkmark$           & $\times$                   \\ \hline
        ~\cite{siblingattack}        &FR                     & Pixel           & $\times$          & $\times$               & $\times$             & $\times$                   \\ 
        ~\cite{advfaces}                   & FR                      & Pixel           & $\times$          & $\times$               & $\times$             & $\times$                   \\ 
        ~\cite{semanticadv}              & FR                       & Semantic              & $\times$          & $\times$               & $\times$             & $\times$                   \\
        ~\cite{GMAA}                          & FR                        & Semantic              & $\times$          & $\times$               & $\times$             & $\times$                   \\ \hline
        ours                      & FR                       & Semantic              & $\times$          & $\checkmark$            & $\checkmark$           & $\checkmark$                 \\ \hline
    \end{tabular}
        }
    \begin{tablenotes}
    \scriptsize
        \item[*] \textbf{FR}: Face Recognition; \textbf{Class}: Classification
        \item[*] \textbf{Pixel}: Pixel-level perturbation; \textbf{Semantic}: Semantic-level perturbation.
        \item[*] \textbf{Combined}: Information hiding-based recovery; \textbf{Self-erasable}: optimization based. 
        \item[*] \textbf{Attack}: Robustness for attack; \textbf{Recovery}: Robustness for recovery.
        \item[*]\textbf{$\checkmark$}: means the scheme achieves this property; \textbf{$\times$}: means it does not.
    \end{tablenotes}
\end{table}
        \subsection{Adversarial Examples}
                Many studies have shown that deep neural networks (DNNs) are highly vulnerable to adversarial examples. Depending on the adversary's knowledge of the target model, these schemes can be categorized into black-box and white-box attacks~\cite{fgsm,pgd}. White-box attacks and query-based black-box attacks~\cite{dong2019efficient}, which rely on knowledge of the target models, are limited in practical scenarios. Therefore, this paper mainly considers transfer-based black-box attacks~\cite{dong2018boosting,xiao2021improving,yang2021towards,zhong2020towards}, which aims at improving transferability without the knowledge of the target model. 
                Although adversarial examples demonstrate extraordinary potential in attacking classification model, the unerasable adversarial performance inevitably prevent their applications in privacy protection scenarios. 

                Recently, a special type of adversarial example, Reversible adversarial examples (RAEs) have been of increasing interest. RAEs can mislead classification model and conduct recovery of origin images, which can server a promising solution for privacy protection.
                Most of them adopt additive perturbations along with reversible data hiding method ~\cite{liu2023unauthorized,xiong2023black,xie2024reversible,yin2019reversible}. They either consider black-box attacks against classification models~\cite{liu2023unauthorized,xiong2023black,zhang2022self}, or a gender protection scenario~\cite{xie2024reversible}. However, there is a gap both in facial privacy protection scenarios, where RAEs cannot support robust recovery that resist image process methods,
                and the self-erasable ability of semantic perturbations.
                
	\subsection{Adversarial Example against Face Recognition}
               Existing attacks against FR systems primarily fall into two categories: patch-based~\cite{sharif2016accessorize,komkov2021advhat} and perturbation-based~\cite{pgd,fgsm,carlini2017towards,dong2019efficient,dong2018boosting,advfaces} attacks. Patch-based attacks, such as Adv-Hat~\cite{komkov2021advhat}, Adv-Makeup~\cite{yin2021adv}, and SOP~\cite{wei2022simultaneously}, add noticeable patches to specific regions of facial images. However, this method significantly degrades image usability. In the other category, perturbation-based adversarial attacks can introduce more imperceptible noise on facial images. 
                 Moreover, most schemes against FR models tend to employ ensemble attacks simply without an in-depth insight in how to improve transferability. Recently, GMAA~\cite{GMAA} improves black-box transferability by introducing a set of the target face with different poses, thus transforming the discrete target domain into a manifold one, which is more likely a data augmentation trick. Sibling-Attack~\cite{siblingattack} uses attacking AR models as an auxiliary task, generating additive adversarial perturbations through multitask optimization. The strategy of Sibling-Attack is enlarging search space, and this additive perturbation will inevitably introduce noticeable noise on facial images. 
                 However, existing schemes lack considerations on how to improve transferability with an in-depth insight into the inherent relationship between FR models. Moreover, all of them cannot support recovery of source facial images, which hinders the authentication and management of TAs.

        \section{Problem Statement}
        In general, there are two kinds of adversarial examples against FR models, including untargeted (dodging) and targeted (impersonation) attacks. 
     Numerous facial privacy protection schemes have been proposed~\cite{hu2022protecting,shamshad2023clip2protect,li2024transferable} aiming at impersonation attacks. 
     As a result, 
    in this section, we formulate a facial privacy protection scenario based on impersonation attacks and provide a comprehensive description of the abilities and goal of users, trusted authority, and unauthorized hackers.
	\label{sec.3}
    \subsection{Users’ Ability and Goal}\label{sec.3.1}
    Users possess several facial images $x_{cov}\in\mathbb{R}^{h\cdot w}$,  where $\mathbb{R}$ denotes the real number field, and $h$, $w$ denotes the height and width of the $x_{cov}$ respectively. Users have no knowledge about the parameters or architecture of malicious FR models but can leverage a locally held facial privacy protection model to process $x_{cov}$, resulting in a protected image $x_{adv}$. In general, the protected face generation process can be expressed as the following formula:
        \begin{equation}
        \begin{aligned}
        \max_{x_{adv}} \quad & \mathcal{L}_{adv} = \mathcal{D}\left(FR(x_{adv}), FR(x_{target})\right) \\
        \text{s.t.} \quad & \|x_{adv} - x_{cov}\|_{p} \leq \delta.
        \end{aligned}
        \label{eq1}
        \end{equation}
    where $\mathcal{D}\left(\cdot\right)$ denotes a distance function. $FR(\cdot)$ denotes a DNN-based FR model. $x_{adv}$ denotes protected faces, $x_{cov}$ denotes source face and $x_{target}$ denotes target face. $||x_{adv},x_{cov}||_{p}\leq\delta$ is utilized to quantify the visual similarity between $x_{adv}$ and $x_{cov}$, where $||\cdot||_{p}$ means $L_p$ norm.
    \subsection{Trusted Authorities’ Ability and Goal}\label{sec.3.2}
    Trusted Authorities (TAs) represent platform managers, and need to conduct face verification in forensics scenarios. TAs can obtain the clean example $x_{rec}$ through the restorer and conduct face verification through the FR model they hold. TAs are totally honest entities and would not deliver restorer and $x_{rec}$ to UHs or other Users. Therefore, we redefine facial privacy protection scenario in erasion perspective. We present a formula for recovery example $x_{rec}$.
        \begin{equation}
        \begin{aligned}
        \min_{x_{rec}} \quad & \mathcal{L}_{erasion} = \mathcal{D}\left(FR(x_{rec}), FR(x_{target})\right) \\
        \text{s.t.} \quad & \|x_{rec} - x_{cov}\|_{p} \leq \delta.
        \end{aligned}
        \label{eq2}
        \end{equation}

    In the process of erasion, we expect the erasion operation to eliminate the adversarial strength towards $x_{target}$ to the greatest extent. We quantify the visual similarity between $x_{rec}$ and $x_{cov}$ through the function $||x_{rec},x_{cov}||_{p}$. 
    \subsection{Unauthorized Hackers’ Ability and Goal}\label{sec.3.3}
    Unauthorized Hackers (UHs) can obtain facial images from online platforms and conduct malicious face verification. 
        Based on their ability to defend against adversarial examples, we define the following two types:
        
        Professional Unauthorized Hackers (\textbf{PUHs}): In this case, PUHs are experts in adversarial examples and face recognition, possessing the ability to fine-tune FR models and resist adversarial threats. 
        
        Regular Unauthorized Hackers (\textbf{RUHs}): In this case, 
        They can only acquire FR models trained on large-scale data from the internet, or use commercial FR models.

\section{Methodology}
	\label{sec.4}
    \begin{figure*}[!tbph]
    \centering
    \scalebox{0.8}{\includegraphics[width=1\textwidth]{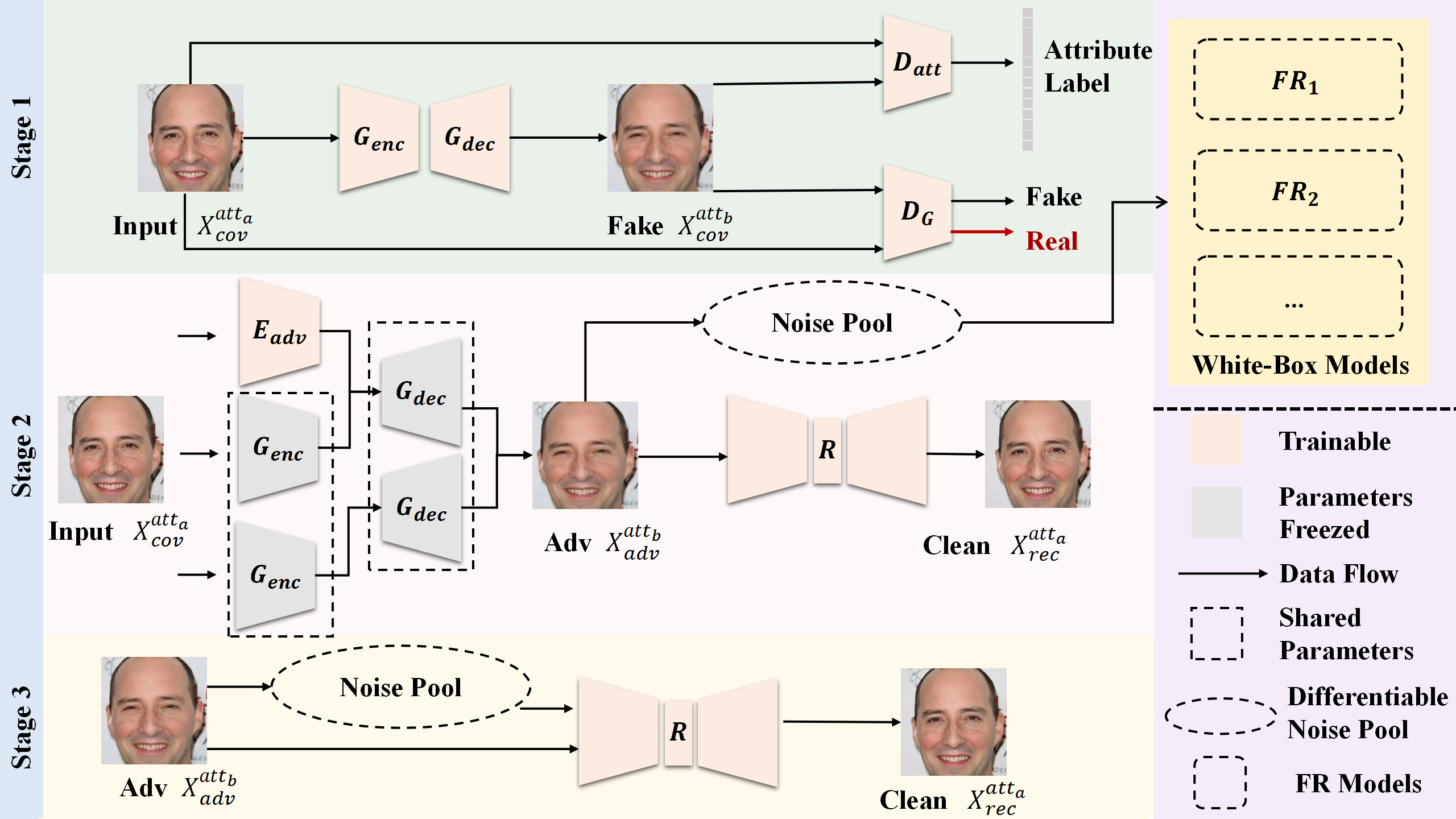}}
    \caption{Pipeline of ErasableMask: A three-stage curriculum learning is introduced to address optimization conflicts between adversarial and erasion performance.
Stage 1: $G_{end}$ and $G_{dec}$ learn an attribute modification strategy.  
Stage 2: $E_{adv}$ and $R$ are tightly coupled and trained end-to-end.  
Stage 3: The robustness and erasion capabilities of $R$ are further strengthened.}
    \label{Fig2}
    \end{figure*}
    As shown in Fig.~\ref{Fig2}, we built the ErasableMask framework using a conditional GAN architecture~\cite{he2019attgan}.  
    ErasableMask consists of a generator $G$, which is composed of an encoder $G_{enc}$, a perturbation encoder $E_{adv}$, and a decoder $G_{dec}$. An restorer $R$, and a discriminator $D$, which has two components including an attribute classifier $D_{att}$ and a discriminator $D_G$.
    \subsection{Visual Identity-preserved Facial Attribute Manipulation}\label{sec.4.2.1}
    As input, the generator $G$ first receives a clean domain facial image $x_{cov}^{att_a}$ with n-bit attribute label $att_a \in \{0,1\}^{n}$ and a n-bit attribute label $att_b \in \{0,1\}^{n}$ to generate a protected example $x_{adv}^{att_b}$. The discriminator $D_G$ learns to distinguish the clean domain image $x_{cov}^{att_a}$ from the protected face $x_{adv}^{att_b}$. Correspondingly, the generator $G$ learns to generate protected faces that match the real image distribution to deceive $D_G$. 
    \begin{equation}
    \centering
    \begin{aligned}
    \mathcal{L}_{D}&=-\mathrm{log~}D_{G}(x_{cov}^{att_{a}})-\mathrm{log~}(1-D_{G}(x_{adv}^{att_{b}})),\\
    \mathcal{L}_{G}&=-\mathrm{log~}(D_{G}(x_{adv}^{att_{b}})).
    \end{aligned}
    \label{eq8}
    \end{equation}

    To ensure that  $x_{adv}^{att_b}$  retains the correct attributes in face, we employ the attribute constraint loss to effectively disentangle facial attribute information in high-dimensional feature space. 
    By constraining  $x_{adv}^{att_b}$ with the $\mathcal{L}_{att}^G$ in attribute classifier $D_{att}$, we reduce the semantic difference between $x_{adv}^{att_b}$  and the target attributes $att_b$. The loss function can be formulated as:

    \begin{equation}
    \centering
    \begin{aligned}
    \mathcal{L}_{att}^D&=\sum_{i=1}^n-att_{a_i}\log D_{att}(x_{cov}^{att_a})\\&-(1-att_{a_i})log (1-D_{att}(x_{cov}^{att_a})),\\
    \mathcal{L}_{att}^G&=\sum_{i=1}^n-att_{b_i}log D_{att}(x_{adv}^{att_b})\\&-(1-att_{b_i})log (1-D_{att}(x_{adv}^{att_b})).
    \end{aligned}
    \label{eq9}
    \end{equation}

    The encoder $G_{enc}$ must ensure that the generated image retains as much of the facial content information from the source image $x_{cov}^{att_a}$ as possible, as the latent variable $z$ needs to enable the decoder $G_{dec}$ to reconstruct attribute-independent details for any attribute conditions. To achieve this, the reconstruction loss $\mathcal{L}_{rec}$ is introduced.
    \begin{equation}
    \centering
    \begin{aligned}
    \mathcal{L}_{rec}=\left\|G_{dec}(G_{enc}(x_{cov}^{att_a}),att_a)-x_{cov}^{att_a}\right\|_1,
    \end{aligned}
    \label{eq10}
    \end{equation}
    where $\|\cdot\|_1$ represents the $L_1$ loss.
\begin{algorithm}[!tbph]
\caption{Meta-auxiliary Attack}
\label{alg1}
\begin{algorithmic}[1]
\REQUIRE

    $x_{cov}^{att_a}$ (source image); $G( \cdot)$ (adversarial example generator);
    $FR_{i}\in \left \{ FR_{1},FR_{2}\dots FR_{K}  \right \}$ (surrogate FR models);
    \\
    
    \ENSURE
    best model parameters;

        \STATE  Initialization: $\mathcal{L}_{FR_i}\{0\} = \mathcal{L}_{FR_i}\{1\} = 1$;    
        \FOR{$i\in  [ 0,Epoch  ]$ }
        \FOR{$x_{cov}^{att_a}\in  Dataset$ }
           \STATE $x_{adv}^{att_b}  \Leftarrow G(x_{cov}^{att_a}, att_b) $;
           
           \FOR{ $p\in [1,\cdots,K]$ }
            \STATE Calculate $\mathcal{L} _{FR_p}^{pri}$ with Eq.~\ref{eq15};
            \STATE $\theta_{E}^{\prime} =\theta_{E}-lr\cdot\nabla_{\theta_{E}} \mathcal{L}_{FR_{A}}^{pri}(\theta_{E})$;
            \FOR{ $q\in \{1,\cdots,K\} \cap q\neq p$ } 
                \STATE $x_{adv}^{\prime att_{b}}  \Leftarrow G\big(x_{cov}^{att_a},\theta_{E}^{\prime},att_{b}\big)$;
                \ENDFOR
                  
                  \STATE Calculate $\mathcal{L} _{FR_p}^{aux}$ with Eq.~\ref{eq17};       
           \ENDFOR

            \STATE Calculate $\mathcal{L}_{adv}$ and update $E_{adv}$ with Eq.~\ref{eq18};
        
           \ENDFOR
           
          \STATE Calculate $w_i(t)$ with Eq.~\ref{eq16};
           
        \ENDFOR
        \RETURN 
        best model parameters;
    \end{algorithmic}
\end{algorithm}
    \subsection{Meta-auxiliary Attack for Semantic Perturbations}\label{sec.4.2.2}
\begin{figure}[!htbp] 
        \centering
        \includegraphics[width=0.5\textwidth]{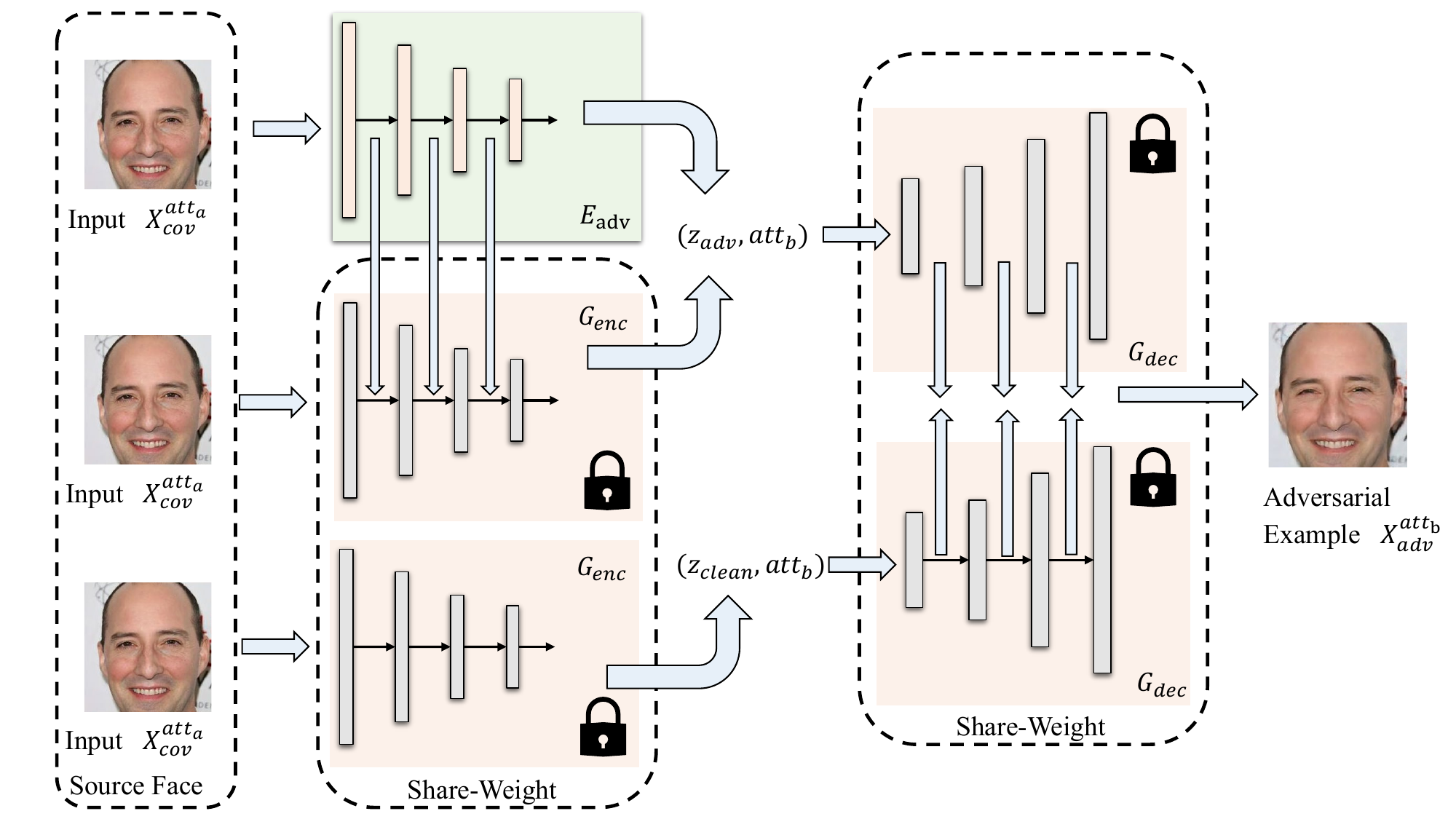}
        \caption{Detailed framework of semantic perturbations and clean-domain information injection.}
        \label{Fig3}
    \end{figure}

    Unlike pixel-level perturbations, ErasableMask achieves a more natural and realistic result for $x_{adv}^{att_b}$ by introducing semantic perturbations, as illustrated in Fig.~\ref{Fig3}. 
    Specifically, $G_{enc}$ and $G_{dec}$ are well-trained under the constraints in section~\ref{sec.4.2.1} to generate naturally realistic attribute-modified images. On this basis, ErasableMask introduces a perturbation encoder $E_{adv}$, which is initially set with parameters identical to $G_{enc}$.
    During training, $E_{adv}$ applies semantic perturbations by fusing its each layer's outputs with each layer of $G_{enc}$'s output. Denoting the $i$-th layer of $G_{enc}$ as $G_{enc}^i$ and the $i$-th layer of $E_{adv}$ as $E_{adv}^i$, where $i \in \{1,\cdots,n\}$ for an encoder with $n$ layers, the operations to the $i$-th layer of $G_{enc}$ can be represented as:
    \begin{equation}
    \centering
    \begin{aligned}
    ft_i&=G_{enc}^i(ft_{i-1}),\\
    perb_i&=E_{adv}^i(perb_{i-1}),\\
    fs_{i}&=\beta \cdot ft_i+(1 - \beta) \cdot perb_i.
    \end{aligned}
    \label{eq13}
    \end{equation}
    
    Specifically, both $G_{enc}$ and $E_{adv}$ take the clean example $x_{cov}^{att_a}$ as their input. The decoding process of $G_{dec}$ is:
    \begin{equation}
    \centering
    \begin{aligned}
    z_{adv}  &= [fs_1,fs_2,\cdots,fs_n],\\
    x_{adv}^{att_b}&=G_{dec}(z_{adv} ,att_b).
     \end{aligned}
    \label{eq14}
    \end{equation}

    To avoid $x_{adv}^{att_b}$ differing from $x_{cov}^{att_{b}}$ largely, a perturbation loss $\mathcal{L}_{perb}$ is introduced.
    \begin{equation}
    \centering
    \begin{aligned}
    \mathcal{L}_{perb}=\max\left(\sigma_{1},\left\|x_{adv}^{att_{b}}-x_{cov}^{att_{b}}\right\|_{2}\right),
    \end{aligned}
    \label{eq11}
    \end{equation}
    where $\sigma_{1} \in (0,+\infty)$, and $\|\cdot\|_2$ represents the $L_2$ norm. In ErasableMask, $\sigma_{1}$ is set to 30.0. 
    Particularly, $x_{cov}^{att_b}$ can be generated from $x_{cov}^{att_b} = G_{dec}(G_{enc}(x_{cov}^{att_a}),att_b)$.
    %

    To enable $x_{adv}^{att_b}$ to demonstrate adversarial performance, we need to employ a transferable attack strategy. However, the existing ensemble attack fails to learn general and robust features due to its unbalanced and unstable learning process. (weak transferability in our experiments).
    %
    Motivated by this, we introduce a more stable and balancing attack strategy called meta-auxiliary attack, to fine-grained fine-tuned the optimization direction of $E_{adv}$ via additional beneficial features from other surrogate models during each attack. The algorithm is demonstrated in Alg.~\ref{alg1}. 
    Specifically, before attacking a specific surrogate FR model, we split $K$ white-box surrogate models $\{FR_i|i\in\{1,\cdots,K\}\}$ into $1$ primary task and $K -1$ auxiliary tasks (All surrogate FR models take turns as primary task). For the convenience of description, we take surrogate model $FR_{A}$ as an example, the rest surrogate models $\{FR_i|{i\in\{1,\cdots,K\}\cap i\neq A}\}$ denote auxiliary tasks. Thereafter, we conduct meta-train in primary task:
    \begin{equation}
    \centering
    \begin{aligned}
    &\mathcal{L}_{FR_A}^{pri}=1-cos\Big[FR_{A}\Big(x_{target}\Big),FR_{A}\Big(x_{adv}^{att_{b}}\Big)\Big].
    \end{aligned}
    \label{eq15}
    \end{equation}

    Then we conduct meta-test in auxiliary tasks to perform fine-grained fine-tuning in optimization direction. 
    \begin{equation}
    \centering
    \small{
    \begin{aligned}
    \theta_{E}^{\prime}&=\theta_{E}-lr\cdot\nabla_{\theta_{E}} \mathcal{L}_{FR_{A}}^{pri}(\theta_{E}),\\
    x_{adv}^{\prime att_{b}}&=G_{dec}\big(G_{enc}(x_{cov}^{att_{a}}),E_{adv}^{\theta_{E}^{\prime}}(x_{cov}^{att_{a}}),att_{b}\big),\\
    \mathcal{L}_{FR_A}^{aux}&=\frac{\sum_{i=1\cap i\neq A}^{K}w_{i}(t)\left(1-cos\left(FR_{i}(x_{target}),FR_{i}(x_{adv}^{\prime att_{b}})\right)\right)}{K-1},
    \end{aligned}}
    \label{eq17}
    \end{equation}
    where we also employ a self-adaptive parameter $w_{i}$ to balance the contributions of different surrogate models. Via this, it can prevent easily attackable surrogate models from dominating during training. We can obtain $w_{i}$ from the following.
    \begin{equation}
    \centering
    \begin{aligned}
    rate_i(t)&=\frac{Mean\left(\mathcal{L}_{FR_i}\{t-1\}\right)}{Mean\left(\mathcal{L}_{FR_i}\{t-2\}\right)},i\in\{1,2\ldots K\},\\
    w_i(t)&=\exp\biggl(\frac{\exp\bigl(rate_i(t)\bigr)}{\sum_i^K\exp\bigl(rate_i(t)\bigr)}\biggr),
    \end{aligned}
    \label{eq16}
    \end{equation}
    where $Mean\left(\mathcal{L}_{FR_i}\{t\}\right)$ is the mean of $\mathcal{L}_{FR_i}$ in t-th epoch.
    Finally, the adversarial loss of ErasableMask is:
    \begin{equation}
    \centering
    \begin{aligned}
    \mathcal{L}_{adv}=\max(\frac1{2K}\sum_i^K(\mathcal{L}_{FR_i}^{pri}+\mathcal{L}_{FR_i}^{aux}),\epsilon),
    \end{aligned}
    \label{eq18}
    \end{equation}
    where $\epsilon$ is used to adjust the perturbation intensity.
    \subsection{Perturbation Erasion based on Information Injection}\label{sec.4.2.3}
    The reconstruction of protected faces is an ill-posed problem, as a single protected face $x_{adv}^{att_b}$ can potentially be reconstructed to numerous $x_{rec}^{att_a}$. Moreover, since the semantic perturbations in feature space permanently damage identifiable information in source faces, this makes it challenging for the restorer $R$ to obtain enough information for perturbation erasion. To address this, ErasableMask injects clean-domain information during generation to maximize the amount of source information needed by $R$.

    Specifically, a copy of $G_{dec}$ is used by introducing an additional branch as shown in Fig.~\ref{Fig3}, where $x_{cov}^{att_a}$ is used as input to obtain the latent variable $ z_{clean} = G_{enc}(x_{cov}^{{att}_a}) $. Subsequently, the tuples $( z_{clean}, {att}_b)$ and $(z_{adv}, att_b)$ are input into $G_{dec}$, where feature fusion is performed after each layer’s output and is controlled by weight $\gamma$. The restorer $R$ is initialized with the same parameters and structures as $G$ and is optimized utilizing the loss $\mathcal{L}_{era}$:
    \begin{equation}
    \centering
    \begin{aligned}
    x_{rec}^{att_a} &= R(x_{adv}^{att_b}), \\
    \mathcal{L}_{era}&=\sum_{n=1}^{N}||{x_{rec}^{att_a}}^{(n)} - {x_{cov}^{att_a}}^{(n)}||_2,
    \end{aligned}
    \label{eq19}
    \end{equation}
    where $N$ is the number of training samples.
    \subsection{Robustness Enhancement and Curriculum Learning}\label{sec.4.2.4}	
    In real-world facial privacy protection scenarios, online social platforms process uploaded images to comply with file storage and transmission standards. This processing significantly affects ErasableMask's adversarial and erasion performance. Therefore, after generating protected faces $x_{adv}^{att_b}$, ErasableMask incorporates a noise pool to enhance its robustness. The noise pool includes JPEG compression ($QF=50$), Gaussian noise ($Var=0.003$), and Resizing transformations ($1/4$).

    In the initial training stage, introducing all losses makes it challenging to learn an effective generation capability. To address this issue, we introduce a curriculum learning strategy that allows the model to gradually acquire generation and perturbation erasion capabilities over three stages. 

    In the first stage, we aim to develop a well-trained encoder $G_{enc}$ and decoder $G_{dec}$. The loss function at this stage is:
            \begin{equation}
            \centering
            \begin{aligned}
            \mathcal{L}_{1}=\lambda_{rec}\mathcal{L}_{rec}+\lambda_{att}\mathcal{L}_{att}^{G}+\lambda_{G}\mathcal{L}_{G}.
            \end{aligned}
            \label{eq20}
            \end{equation}
            
    In the second stage, we aim to ensure that $\mathcal{L}_{adv}$ does not impair the model’s ability to produce natural images. Thus, in this phase, we fix the parameters of the encoder $G_{enc}$ and decoder $G_{dec}$, then introduce $E_{adv}$, which has the same structure and is initialized with the parameters of $G_{enc}$, to generate semantic perturbations. Additionally, we introduce $R$ at this stage, initializing it with the parameters of the encoder $G_{enc}$ and decoder $G_{dec}$. This stage allows $E_{adv}$ and the $R$ to be tightly coupled and can be trained end-to-end.
    The loss function of stage 2 is:
            \begin{equation}
            \centering
            \begin{aligned}
            \mathcal{L}_{2}=\lambda_{att}\mathcal{L}_{att}^{G}+\lambda_{rec}\mathcal{L}_{rec}+\lambda_{G}\mathcal{L}_{G}\\+\lambda_{adv}\mathcal{L}_{adv}+\lambda_{era}\mathcal{L}_{era}+\lambda_{perb}\mathcal{L}_{perb}. %
            \end{aligned}
            \label{eq21}
            \end{equation}
            
    Since erasing semantic perturbations and obtaining a non-adversarial example $x_{rec}^{att_a}$ from $x_{adv}^{att_b}$ is a challenging task, the third stage focuses on independently training the $R$ to develop its ability to map from the adversarial domain to the clean domain. The loss function of stage 3 is:
            \begin{equation}
            \centering
            \begin{aligned}
            \mathcal{L}_3=\mathcal{L}_{era}.
            \end{aligned}
            \label{eq22}
            \end{equation}
	\section{Experiment} \label{sec.5}
\begin{table*}[!tbph]
 \caption{ASR and ESR Results against Black-Box FR model. 
 }
\label{table_2}
	\renewcommand\arraystretch{2.0}  
    \centering
\resizebox{1.0\textwidth}{!}{
\fontsize{15}{10}\selectfont
       \begin{tabular}{c|c|c|cccccccccccc}
       \hline
           \multirow{2}{*}{\textbf{Datasets}} &\multirow{2}{*}{\textbf{Type}} & \multirow{2}{*}{\textbf{Scheme}}&\multicolumn{2}{c}{\textbf{Facenet}} &\multicolumn{2}{c}{\textbf{ArcFace}}&\multicolumn{2}{c}{\textbf{IRSE50}} &\multicolumn{2}{c}{\textbf{MobileFace}} & \multicolumn{2}{c}{\textbf{CosFace}} &\multicolumn{2}{c}{\textbf{IR152}} 
            \\ \cline{4-15}
			&&& ASR↑&ESR↑& ASR↑&ESR↑& ASR↑&ESR↑& ASR↑&ESR↑& ASR↑&ESR↑& ASR↑&ESR↑ \\
            \hline
		\multirow{9}{*}{CelebA-HQ}&\multirow{2}{*}{Without adversarial perturbation}
     &Clean & 0.45 & - & 0.0 &-&0.50 &- &0.05 &- & 0.0 &-  &1.65&- \\
		& &Attributes Modified&0.22  &- &0.0 &- &0.0 &- &0.0  &- &0.0 &-  &1.7  &- \\
			\cline{2-15}
             &\multirow{3}{*}{Gradient-based}
             &FGSM~\cite{fgsm}&2.27 &- &35.00 &-&61.36  &- & 31.82 &- &33.18 &- &42.27 & - \\
			 && PGD~\cite{pgd}&3.63 & - & 65.45 &- & 75.45 &- &45.46&- & 64.54 &-& 59.09 &- \\
             &&Sibling Attack~\cite{siblingattack}&14.93 & - & \textbf{93.21} &- & \textbf{95.48} &- &83.25&- & \underline{92.31} &- & \textbf{98.19} &- \\
			
              \cline{2-15}
               &\multirow{3}{*}{Face-based}
               &AdvFaces~\cite{advfaces} &\underline{35.00} &- &50.00 &-&83.67  &- &\underline{91.33}  &- &78.67 &-  &84.66 &-  \\
			 &&SemanticAdv~\cite{semanticadv}&3.67 &-  &10.66  &-&38.24  &- &47.79 &-  &48.16  &- &63.24 &- \\
             &&GMAA~\cite{GMAA}&16.04 &-  &44.79  &-&59.91  &- &13.85 &-  &50.59  &- &61.29 &- \\	
              \cline{2-15}
             & Erasable face-based &\textbf{ ErasableMask (ours)} &\textbf{64.00} &96.67 & \underline{90.44} &98.22&\underline{94.00} &89.33 &\textbf{92.22} &94.89 & \textbf{93.33} &98.89  &\underline{92.00} &88.00\\
			\hline\hline

   				\multirow{9}{*}{FFHQ}&\multirow{2}{*}{Without adversarial perturbation}
     &Clean & 0.0 & - & 0.0 &-&0.0 &- &0.0 &- & 0.0 &-  &2.0&- \\
		& &Attributes Modified&0.0  &- &0.0 &- &0.0 &- &0.0  &- &0.0 &-  &1.11  &- \\
			\cline{2-15}
             &\multirow{3}{*}{Gradient-based}
             &FGSM~\cite{fgsm}&2.17 &- &38.41 &-&47.83  &- & 23.19 &- &32.61 &- &52.89 & - \\
			 && PGD~\cite{pgd}&5.04 & - & 68.34 &- & 62.58 &- &34.53&- & 63.31 &-& 65.46 &- \\
             &&Sibling Attack~\cite{siblingattack}&15.94 & - & \textbf{83.33} &- & \underline{88.41} &- &66.67&- & \underline{85.51} &- & \underline{89.86} &- \\
			
              \cline{2-15}
               &\multirow{3}{*}{Face-based}
               &AdvFaces~\cite{advfaces} &\underline{25.60} &- &46.80 &-&77.40  &- &\underline{68.80}  &- &38.20 &-  &67.80 &-  \\
			 &&SemanticAdv~\cite{semanticadv}&5.88 &-  &21.33  &-&31.98  &- &15.07 &-  &22.06  &- &35.66 &- \\
             &&GMAA~\cite{GMAA}&15.94 &-  &21.49  &-&46.01 &- &11.71 &-  &21.74  &- &38.53 &- \\	
              \cline{2-15}
             & Erasable face-based &\textbf{ ErasableMask (ours)} &\textbf{58.22} &98.22 & \underline{81.78} &99.11&\textbf{97.33} &93.78&\textbf{93.11} &98.22 & \textbf{92.67} &99.56  &\textbf{90.67} &89.33\\
        \hline
	\end{tabular}}

\end{table*}

        \subsection{Experimental setting}
        \subsubsection{Implementation details and Dataset}
                    To maintain all losses on the same scale, we set the parameters $\lambda_{att}$, $\lambda_{rec}$, $\lambda_{G}$, $\lambda_{adv}$, $\lambda_{era}$, and $\lambda_{perb}$ to $10$, $150$, $1$, $200$, $150$, and $1$ respectively. ErasableMask is trained with 200 epochs in the first stage, 100 epochs in the second stage, and 50 epochs in the third stage. The learning rate is set to 0.00002.
                    We selected high-resolution datasets: CelebA-HQ~\cite{karras2017progressive}, a high-quality facial image dataset containing approximately 30,000 images with a resolution of $512\times512$, and FFHQ~\cite{karras2019style}, a high-quality face dataset with over 70,000 images. To simulate real-world applications, we first use MTCNN~\cite{zhang2016joint} to extract and preprocess the facial regions within images. Due to computational constraints, each image in the dataset is downscale to a resolution of $256\times256$. All experiments are conducted on RTX3090 GPU 24 GB$\times$1.
                \subsubsection{Practical Evaluation Metrics for Facial Privacy Protection}\label{sec.3.3}
                We use \textbf{Attack Success Rate (ASR)} to evaluate the ability of facial privacy protection.
                \begin{equation}
                \centering
                \begin{aligned}
            ASR=\frac{\sum_i^N1\left(\cos[FR(x_{adv}),FR(x_{target})]>\tau_1\right)}N.
                \end{aligned}
                \label{eq3}
                \end{equation}
                Where, $1\left( \cdot \right)$ denotes indicator function, ASR indicates that facial privacy protection is considered successful when the similarity between $x_{adv}$ and $x_{target}$ exceeds the threshold $\tau_1$. In this paper, $\tau_1$ was set to 0.01 times the False Acceptance Rate (FAR) of the FR model.
            
                To evaluate ErasableMask's performance of erasion, we propose the \textbf{Erasion Success Rate (ESR)} as a metric. 
                \begin{equation}
                \centering
                \begin{aligned}
            ESR=\frac{\sum_i^N1\left(\cos[FR(x_{rec}),FR(x_{target})]<\tau_2\right)}N.
                \end{aligned}
                \label{eq4}
                \end{equation}
                ESR indicates that ErasableMask has successfully reconstructed a clean example that does not disrupt face verification when the cosine similarity between $x_{rec}$ and $x_{target}$ is less than 0.1 times the False Acceptance Rate (FAR) of the FR model.
        \subsubsection{Competitors}
                    To verify the performance of ErasableMask's black-box transferability, we implemented several benchmark adversarial attack schemes, including FGSM~\cite{fgsm}, PGD~\cite{pgd}, Sibling Attack~\cite{siblingattack}, AdvFaces~\cite{advfaces}, Semanticadv~\cite{semanticadv}, and GMAA~\cite{GMAA}. FGSM and PGD are well-known schemes due to their strong adversarial capabilities, and we set the perturbation strength to $4/255$, considering noticeable noise introduced by additive perturbations. 
                    AdvFaces, Sibling Attack, Semanticadv and GMAA are recent schemes that generate adversarial examples specifically targeting FR models.
        \subsubsection{Target models}
                    Following~\cite{hu2022protecting}, we conducted extensive experiments and selected six FR models as white-box models: IR152~\cite{he2016deep}, IRSE50~\cite{hu2018squeeze}, Facenet~\cite{schroff2015facenet}, Mobileface~\cite{deng2019arcface}, ArcFace~\cite{deng2019arcface} based on IResNet100~\cite{duta2021improved}, and CosFace~\cite{wang2018cosface} based on IResNet100. Three of these models were used as white-box models for training, while the remaining three were used as black-box models for evaluation. Subsequently, Face++~\cite{MEGVII}, Tencent~\cite{tencentapi} and Aliyun~\cite{Aliyun} were chosen as black-box commercial FR models to serve as target models. To evaluate transferability against black-box robust FR models, we fine-tuned ArcFace and CosFace on the public dataset LFW using adversarial training~\cite{madry2017towards}, achieving $99.28\%$ and $99.35\%$ accuracy on LFW dataset, respectively. 

        \subsection{Comparison on Offline FR models}\label{blackbox}
                        We benchmark against prior work~\cite{hu2022protecting,shamshad2023clip2protect,GMAA} in Table.~\ref{table_2}, concentrating on evaluating ErasableMask's transferability against black-box FR models. ErasableMask almost achieves optimal ASR among all competitors, surpassing all face-based schemes. Different from existing schemes, ErasableMask also shows strong transferability against FaceNet, with $30\%$ higher than the second optimal. 

                         Compared with existing facial protection schemes, ErasableMask offers a restorer that can perform perturbation erasion and obtain a clean-domain example. The perturbation erasion performance is also demonstrated in Table.~\ref{table_2}. The ESR results against all black-box FR models are almost close to $90\%$. 
        \subsection{Comparison on Commercial FR systems}\label{comercial api}
        \begin{table}[!t]
        \renewcommand{\arraystretch}{2.0}
                \centering
                \caption{The mean confidence returned from commercial FR systems.
                }
                \resizebox{0.5\textwidth}{!}{
                \fontsize{25}{20}\selectfont
                \begin{tabular}{ccccccccc}  
                \hline
                \textbf{Scheme}     & \multicolumn{2}{c}{\textbf{Aliyun Confidence}}                                            & \multicolumn{2}{c}{\textbf{Face++ Confidence}}                                            & \multicolumn{2}{c}{\textbf{Tencent Confidence}}                                           & \multicolumn{2}{c}{\textbf{Mean Confidence}}                                   \\ \hline
Metric              & \multicolumn{1}{l}{Attack ↑} & \multicolumn{1}{l}{Erasion ↓} & \multicolumn{1}{l}{Attack ↑} & \multicolumn{1}{l}{Erasion ↓} & \multicolumn{1}{l}{Attack ↑} & \multicolumn{1}{l}{Erasion ↓} & \multicolumn{1}{l}{Attack ↑} & \multicolumn{1}{l}{Erasion ↓} \\ \hline
Clean               & 4.65                                  & -                                      & 39.92                                 & -                                      & 8.66                                  & -                                      & 17.74                                 & -                                      \\
FGSM                & 29.09                                 & -                                      & 52.41                                 & -                                      & 24.34                                 & -                                      & 35.28                                 & -                                      \\
PGD                 & 31.99                                 & -                                      & 56.57                                 & -                                      & 26.07                                 & -                                      & 38.21                                 & -                                      \\
Sibling Attack      & 50.63                                 & -                                      & 67.39                                 & -                                      & 42.71                                 & -                                      & 53.57                                 & -                                      \\
AdvFaces            & 43.38                                 & -                                      & \underline{70.65}        & -                                      & \underline{47.72 }       & -                                      & 53.92                                 & -                                      \\
SemanticAdv         & 30.19                                 & -                                      & 55.05                                 & -                                      & 29.12                                 & -                                      & 38.12                                 & -                                      \\
GMAA                & \underline{58.22}        & -                                      & 66.82                                 & -                                      & 45.68                                 & -                                      & \underline{56.91}        & -                                      \\ \hline
ErasableMask (ours) & \textbf{67.01}                        & 5.11                                   & \textbf{78.19}                        & 42.72                                  & \textbf{69.06}                        & 10.15                                  & \textbf{72.60}                        & 23.92                            \\
                \hline
                \end{tabular}
                }
                \label{table_5}
\begin{tablenotes}
    \scriptsize
        \item[*] For Attack, the higher (↑) the similarity confidence score between protected faces and target face returned by commercial systems, the better. 
        \item[*] For Erasion, the lower (↓) the similarity confidence score between protected faces and target face returned by commercial systems, the better. 
\end{tablenotes}
    
            \end{table}
                        Another scenario is that unauthorized hackers choose commercial FR systems, like Aliyun, Tencent, and Face++. To simulate this, we employ commercial FR systems for face verification. We randomly select 100 examples for each scheme, then upload them and target face to platforms. We calculate the mean confidence rate and the results are demonstrated in Table.~\ref{table_5}. The mean confidence of ErasableMask is $9.79\%$, $7.54\%$, and $21.34\%$ higher than the second higher scheme against Aliyun, Face++ and Tencent, respectively.

        \subsection{Comparison on Robust FR Models}
        \begin{table}[]
        \renewcommand{\arraystretch}{1.2}
        \caption{The ASR results and ESR result against robust FR models.}
                \label{table_8}
                \resizebox{0.5\textwidth}{!}{
\begin{tabular}{cccccccccc}
\hline
\multicolumn{2}{c}{\textbf{Black-Box}}                                        & \multicolumn{4}{c}{\textbf{Robust ArcFace}}                             & \multicolumn{4}{c}{\textbf{Robust CosFace}}                             \\ \hline
\multicolumn{2}{c}{Metric}                                                    & \multicolumn{2}{c}{ASR ↑}            & \multicolumn{2}{c}{ESR ↑}            & \multicolumn{2}{c}{ASR ↑}            & \multicolumn{2}{c}{ESR ↑}            \\\hline
\multicolumn{2}{c}{Clean}                                                     & \multicolumn{2}{c}{0.0}            & \multicolumn{2}{c}{-}              & \multicolumn{2}{c}{0.0}            & \multicolumn{2}{c}{-}              \\
\multicolumn{2}{c}{FGSM~\cite{fgsm}}                    & \multicolumn{2}{c}{2.73}           & \multicolumn{2}{c}{-}              & \multicolumn{2}{c}{1.36}           & \multicolumn{2}{c}{-}              \\
\multicolumn{2}{c}{PGD~\cite{pgd}}                      & \multicolumn{2}{c}{5.45}           & \multicolumn{2}{c}{-}              & \multicolumn{2}{c}{3.64}           & \multicolumn{2}{c}{-}              \\
\multicolumn{2}{c}{Sibling Attack~\cite{siblingattack}} & \multicolumn{2}{c}{30.32}          & \multicolumn{2}{c}{-}              & \multicolumn{2}{c}{25.34}          & \multicolumn{2}{c}{-}              \\
\multicolumn{2}{c}{AdvFaces~\cite{advfaces}}            & \multicolumn{2}{c}{36.0}           & \multicolumn{2}{c}{-}              & \multicolumn{2}{c}{61.0}           & \multicolumn{2}{c}{-}              \\
\multicolumn{2}{c}{SemanticAdv~\cite{semanticadv}}      & \multicolumn{2}{c}{0.36}           & \multicolumn{2}{c}{-}              & \multicolumn{2}{c}{11.76}          & \multicolumn{2}{c}{-}              \\
\multicolumn{2}{c}{GMAA~\cite{GMAA} }                    & \multicolumn{2}{c}{5.54}           & \multicolumn{2}{c}{-}              & \multicolumn{2}{c}{9.90}           & \multicolumn{2}{c}{-}              \\ \hline
\multicolumn{2}{c}{ErasableMask (ours)}                                       & \multicolumn{2}{c}{\textbf{79.78}} & \multicolumn{2}{c}{\textbf{100.0}} & \multicolumn{2}{c}{\textbf{81.33}} & \multicolumn{2}{c}{\textbf{98.44}}  \\              \hline
\end{tabular}                }
\end{table}
                        In this section, we evaluate all competitors and ErasableMask against fine-tuned robust FR models. The results are demonstrated in Table.~\ref{table_8}. ErasableMask outperforms all competitors with $79.78\%$ against ArcFace and $81.33\%$ against CosFace, respectively. Traditional gradient-based schemes exhibit the poorest performance against fine-tuned FR models. However, Sibling Attack employing attacking AR model as an auxiliary task exhibits higher robust performance than other gradient-based schemes. Among all face-based competitors, AdvFaces shows the highest ASR, mainly because AdvFaces generates additive perturbations rather than semantic perturbations. That is to say, semantic perturbation results in a poorer robustness compared with additive perturbation. However, ErasableMask overcomes the drawbacks of semantic perturbation, with over $70\%$ higher than others.
\begin{table*}[!tbph]
\renewcommand{\arraystretch}{2.0}
\centering
\caption{ASR and ESR results after various image processing operations.}
\label{robustness_fig}
\resizebox{0.9\textwidth}{!}{
\fontsize{20}{12}\selectfont
\begin{tabular}{c|c|c|cc|cc|cc|cc|cc|cc} 
\hline\hline
\multirow{2}{*}{\textbf{Processing}} & \multirow{2}{*}{\textbf{Type}}  & \multirow{2}{*}{\textbf{Scheme}} & \multicolumn{2}{c}{\textbf{FaceNet}} & \multicolumn{2}{c}{\textbf{ArcFace}} & \multicolumn{2}{c}{\textbf{IRSE50}} & \multicolumn{2}{c}{\textbf{MobileFace}} & \multicolumn{2}{c}{\textbf{CosFace}} & \multicolumn{2}{c}{\textbf{IR152}}  \\ 
\cline{4-15}
                                     &                                 &                                  & ASR ↑           & ESR ↑                  & ASR ↑            & ESR ↑                 & ASR ↑           & ESR ↑                 & ASR ↑           & ESR ↑                     & ASR ↑          & ESR ↑                  & ASR ↑           & ESR ↑                 \\ 
\hline
\multirow{7}{*}{Resize}              & \multirow{3}{*}{Gradient-based} & FGSM~\cite{fgsm}                             & 1.4           & -                    & 18.6           & -                   & 32.7          & -                   & 14.1          & -                       & 7.3           & -                    & 22.7          & -                   \\
                                     &                                 & PGD~\cite{pgd}                              & 1.8           & -                    & 26.4           & -                   & 34.5          & -                   & 19.1          & -                       & 9.1           & -                    & 26.4          & -                   \\
                                     &                                 & Sibling Attack~\cite{siblingattack}                   & 12.2          & -                    & 59.3           & -                   & 85.1          & -                   & 66.9          & -                       & 39.4          & -                    & \underline{81.9}  & -                   \\ 
\cline{2-15}
                                     & \multirow{3}{*}{Face-based}     & AdvFaces~\cite{advfaces}                         & \underline{34.7}  & -                    & \underline{62.0}   & -                   & \underline{88.0}  & -                   & \underline{80.3}  & -                       & \underline{48.0}  & -                    & 65.0          & -                   \\
                                     &                                 & SemanticAdv~\cite{semanticadv}                      & 2.9           & -                    & 9.6            & -                   & 37.1          & -                   & 45.6          & -                       & 44.5          & -                    & 61.4          & -                   \\
                                     &                                 & GMAA~\cite{GMAA}                             & 8.2           & -                    & 18.4           & -                   & 48.1          & -                   & 8.4           & -                       & 31.3          & -                    & 47.3          & -                   \\ 
\cline{2-15}
                                     & Erasable face-based             & \textbf{ ErasableMask (ours)}    & \textbf{63.6} & 92.7                 & \textbf{90.4}  & 95.8                & \textbf{93.6} & 82.4                & \textbf{92.7} & 84.4                    & \textbf{92.4} & 96.0                 & \textbf{92.2} & 94.2                \\ 
\hline\hline
\multirow{7}{*}{JPEG}                & \multirow{3}{*}{Gradient-based} & FGSM~\cite{fgsm}                             & 2.3           & -                    & 16.8           & -                   & 35.9          & -                   & 14.1          & -                       & 7.3           & -                    & 20.9          & -                   \\
                                     &                                 & PGD~\cite{pgd}                              & 3.6           & -                    & 24.5           & -                   & 38.2          & -                   & 20.9          & -                       & 10.9          & -                    & 23.6          & -                   \\
                                     &                                 & Sibling Attack~\cite{siblingattack}                   & 14.0          & -                    & \underline{71.5}   & -                   & \underline{91.9}  & -                   & 71.9          & -                       & \underline{57.0}  & -                    & \underline{89.6}  & -                   \\ 
\cline{2-15}
                                     & \multirow{3}{*}{Face-based}     & AdvFaces~\cite{advfaces}                         & \underline{33.7}  & -                    & 45.3           & -                   & 72.7          & -                   & \underline{82.0}  & -                       & 47.0          & -                    & 69.3          & -                   \\
                                     &                                 & SemanticAdv~\cite{semanticadv}                      & 4.0           & -                    & 4.0            & -                   & 36.4          & -                   & 42.3          & -                       & 24.6          & -                    & 53.7          & -                   \\
                                     &                                 & GMAA~\cite{GMAA}                             & 12.9          & -                    & 26.8           & -                   & 49.1          & -                   & 9.9           & -                       & 37.0          & -                    & 55.4          & -                   \\ 
\cline{2-15}
                                     & Erasable face-based             & \textbf{ ErasableMask (ours)}    & \textbf{63.8} & 91.3                 & \textbf{90.2}  & 95.8                & \textbf{93.6} & 81.8                & \textbf{92.7} & 86.0                    & \textbf{92.9} & 93.8                 & \textbf{91.6} & 76.4                \\ 
\hline\hline
\multirow{7}{*}{Gaussian noise}      & \multirow{3}{*}{Gradient-based} & FGSM~\cite{fgsm}                             & 2.3           & -                    & 23.6           & -                   & 53.2          & -                   & 20.5          & -                       & 17.7          & -                    & 33.2          & -                   \\
                                     &                                 & PGD~\cite{pgd}                              & 3.6           & -                    & 41.8           & -                   & 62.7          & -                   & 28.2          & -                       & 29.1          & -                    & 46.4          & -                   \\
                                     &                                 & Sibling Attack~\cite{siblingattack}                   & 14.5          & -                    & \underline{87.3}   & -                   & \textbf{95.0} & -                   & 79.2          & -                       & \underline{85.9}  & -                    & \textbf{97.7} & -                   \\ 
\cline{2-15}
                                     & \multirow{3}{*}{Face-based}     & AdvFaces~\cite{advfaces}                         & \underline{31.3}  & -                    & 47.3           & -                   & 81.7          & -                   & \underline{87.6}  & -                       & 71.6          & -                    & 80.7          & -                   \\
                                     &                                 & SemanticAdv~\cite{semanticadv}                      & 2.6           & -                    & 2.9            & -                   & 29.0          & -                   & 26.8          & -                       & 28.3          & -                    & 47.4          & -                   \\
                                     &                                 & GMAA~\cite{GMAA}                             & 12.6          & -                    & 21.3           & -                   & 50.3          & -                   & 6.8           & -                       & 31.2          & -                    & 50.3          & -                   \\ 
\cline{2-15}
                                     & Erasable face-based             & \textbf{ ErasableMask (ours)}    & \textbf{63.7} & 93.1                 & \textbf{88.2}  & 96.0                & \underline{92.9}  & 80.7                & \textbf{91.6} & 86.9                    & \textbf{89.3} & 94.9                 & \underline{91.1}  & 81.6                \\ 
\hline\hline
\multirow{7}{*}{Median Filter}       & \multirow{3}{*}{Gradient-based} & FGSM~\cite{fgsm}                             & 1.8           & -                    & 18.6           & -                   & 37.3          & -                   & 15.5          & -                       & 6.4           & -                    & 22.3          & -                   \\
                                     &                                 & PGD~\cite{pgd}                              & 2.7           & -                    & 29.1           & -                   & 41.8          & -                   & 21.8          & -                       & 9.1           & -                    & 23.6          & -                   \\
                                     &                                 & Sibling Attack~\cite{siblingattack}                   & 13.6          & -                    & \underline{59.7}   & -                   & 83.7          & -                   & 64.3          & -                       & 38.5          & -                    & \underline{83.7}  & -                   \\ 
\cline{2-15}
                                     & \multirow{3}{*}{Face-based}     & AdvFaces~\cite{advfaces}                         & \underline{35.3 } & -                    & 59.3           & -                   & \underline{87.3}  & -                   & \underline{74.0}  & -                       & 31.0          & -                    & 55.0          & -                   \\
                                     &                                 & SemanticAdv~\cite{semanticadv}                      & 1.5           & -                    & 7.7            & -                   & 38.2          & -                   & 43.0          & -                       & 43.0          & -                    & 58.8          & -                   \\
                                     &                                 & GMAA~\cite{GMAA}                             & 15.8          & -                    & 47.5           & -                   & 56.4          & -                   & 7.9           & -                       & \underline{45.5}  & -                    & 60.4          & -                   \\ 
\cline{2-15}
                                     & Erasable face-based             & \textbf{ErasableMask (ours)}     & \textbf{63.3} & 92.7                 & \textbf{90.4}  & 96.0                & \textbf{93.1} & 83.8                & \textbf{92.7} & 86.7                    & \textbf{92.9} & 96.0                 & \textbf{92.4} & 78.7                \\ 
\hline\hline
\multirow{7}{*}{Random Rotate}       & \multirow{3}{*}{Gradient-based} & FGSM~\cite{fgsm}                             & 0.9           & -                    & 9.1            & -                   & 12.7          & -                   & 6.4           & -                       & 3.6           & -                    & 13.2          & -                   \\
                                     &                                 & PGD~\cite{pgd}                              & 2.7           & -                    & 10.9           & -                   & 16.4          & -                   & 7.3           & -                       & 5.5           & -                    & 20.0          & -                   \\
                                     &                                 & Sibling Attack~\cite{siblingattack}                   & 11.3          & -                    & 31.7           & -                   & 41.2          & -                   & 23.9          & -                       & 20.8          & -                    & \underline{52.5}  & -                   \\ 
\cline{2-15}
                                     & \multirow{3}{*}{Face-based}     & AdvFaces~\cite{advfaces}                         & \underline{27.3}  & -                    & \underline{44.3}   & -                   & \underline{57.0}  & -                   & \underline{35.7}  & -                       & 33.7          & -                    & 46.7          & -                   \\
                                     &                                 & SemanticAdv~\cite{semanticadv}                      & 2.6           & -                    & 4.8            & -                   & 15.1          & -                   & 14.3          & -                       & 30.9          & -                    & 40.4          & -                   \\
                                     &                                 & GMAA~\cite{GMAA}                             & 10.9          & -                    & 30.7           & -                   & 39.6          & -                   & 4.9           & -                       & \underline{43.6}  & -                    & 48.5          & -                   \\ 
\cline{2-15}
                                     & Erasable face-based             & \textbf{ ErasableMask (ours)}    & \textbf{63.6} & 42.9                 & \textbf{82.7}  & 36.2                & \textbf{68.2} & 36.7                & \textbf{47.1} & 67.3                    & \textbf{86.4} & 23.8                 & \textbf{81.6} & 28.7                \\ 
\hline\hline
\multirow{7}{*}{Central Crop}        & \multirow{3}{*}{Gradient-based} & FGSM~\cite{fgsm}                             & 2.7           & -                    & 12.7           & -                   & 34.1          & -                   & 9.5           & -                       & 4.5           & -                    & 22.7          & -                   \\
                                     &                                 & PGD~\cite{pgd}                              & 2.7           & -                    & 21.8           & -                   & 37.3          & -                   & 14.5          & -                       & 8.2           & -                    & 26.4          & -                   \\
                                     &                                 & Sibling Attack~\cite{siblingattack}                   & 14.9          & -                    & 47.9           & -                   & 84.2          & -                   & 59.3          & -                       & 30.8          & -                    & \underline{76.0}  & -                   \\ 
\cline{2-15}
                                     & \multirow{3}{*}{Face-based}     & AdvFaces~\cite{advfaces}                         & \underline{38.3 } & -                    & \underline{58.3}   & -                   & \underline{88.0}  & -                   & \underline{76.7}  & -                       & \underline{44.0}  & -                    & 65.3          & -                   \\
                                     &                                 & SemanticAdv~\cite{semanticadv}                      & 2.9           & -                    & 9.2            & -                   & 37.1          & -                   & 45.2          & -                       & 45.6          & -                    & 60.3          & -                   \\
                                     &                                 & GMAA~\cite{GMAA}                             & 15.8          & -                    & 44.6           & -                   & 60.4          & -                   & 11.9          & -                       & 49.5          & -                    & 62.4          & -                   \\ 
\cline{2-15}
                                     & Erasable face-based             & \textbf{ ErasableMask (ours)}    & \textbf{67.6} & 44.0                 & \textbf{92.0 } & 18.9                & \textbf{94.2} & 4.9                 & \textbf{90.0} & 14.0                    & \textbf{92.0} & 14.9                 & \textbf{90.2} & 12.0                \\
\hline\hline
\end{tabular}
}
\end{table*}
        \subsection{Robustness Analysis}
                    
                        We evaluate robustness via applying JPEG($QF=50$), Gaussian noise($Var=0.003$), Resizing transformation($1/2$), Median Filter($Kernel = 5$), Random Rotate($angle \in [-30,30]$), and Central Crop($224$) to the protected faces in Table.~\ref{robustness_fig}. ErasableMask demonstrates excellent protection and erasion capability in terms of ASR and ESR across.
                        It outperforms almost all competitors in both white-box and black-box image processing methods, with rather high ASR. While it also shows excellent erasion capability in JPEG, Gaussian noise, Resizing transformation, Median Filter. Even though the erasion performance against Random Rotate and Central Crop shows less effectiveness, we think it is reasonable that TAs will not apply these methods to disrupt face verification.
        \subsection{Visualization and Image Quality}

\begin{figure}[!tbph]
            \renewcommand{\arraystretch}{1.0}
            \centering
            \scalebox{0.95}{\includegraphics[width=0.5\textwidth]{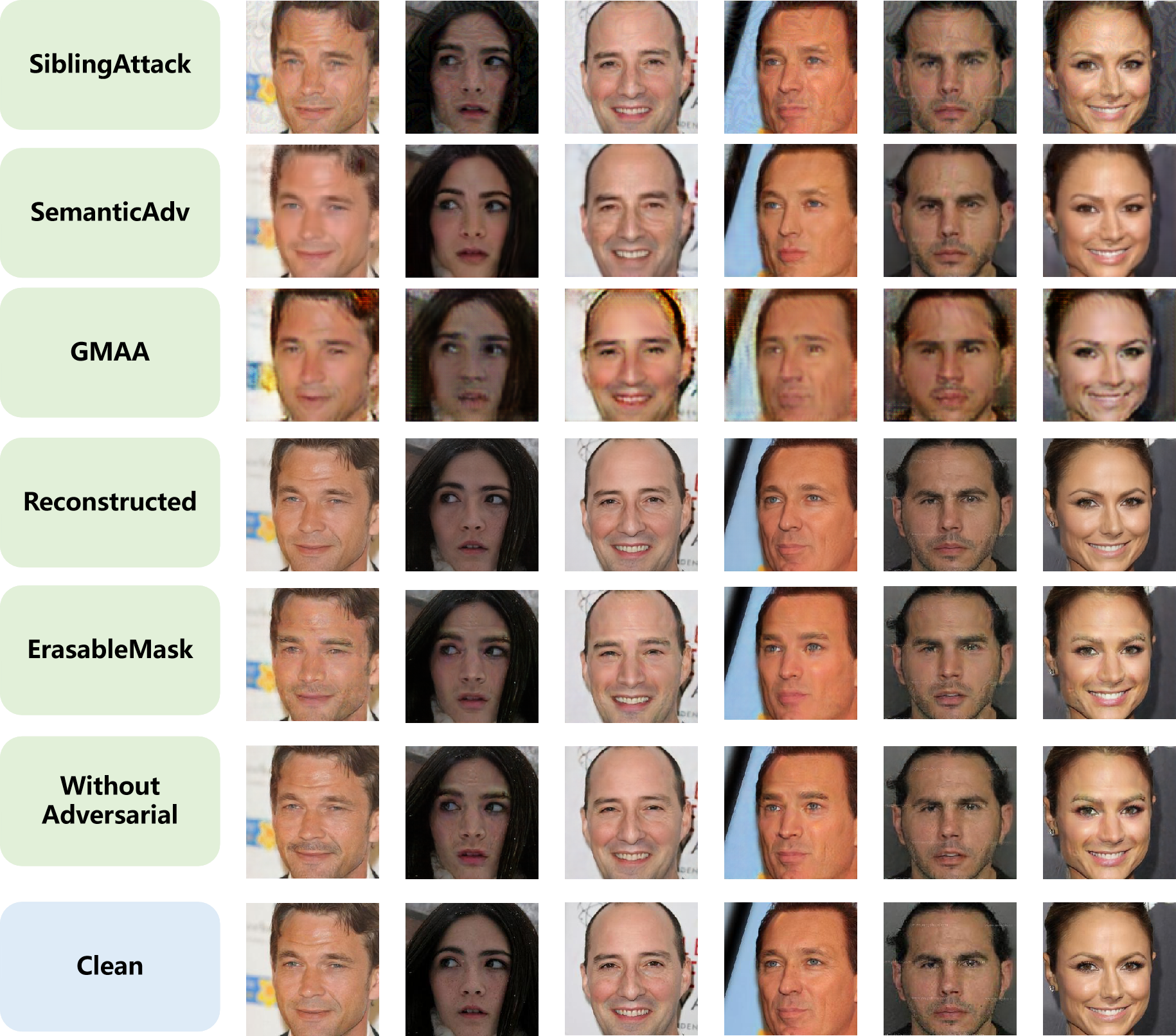}}
            \caption{Visualization of various schemes.  }
            \label{Fig_visual}
            \end{figure}

                        Fig.~\ref{Fig_visual} demonstrates the protected faces generated by various schemes. Compared with gradient-based schemes, ours has no obvious pixel-level adversarial noise.
                        Furthermore, we use L1 loss, Mean Square Error (MSE), Frechet Inception Distance (FID)~\cite{heusel2018ganstrainedtimescaleupdate} and Learned Perceptual Image Similarity(LPIPS)~\cite{zhang2018unreasonableeffectivenessdeepfeatures} as metrics to evaluate the quality of the protected faces. 
                        we compare our protected faces with attribute-modified images without adversarial perturbations for fair comparison. The comparison results are shown in Table.~\ref{quality}. the results indicate that the quality of ErasableMask is similar or less than other schemes with much higher transferability. 
                    \begin{table}
                        \centering
                        \caption{The image quality evaluation of different schemes }
                        \resizebox{0.5\textwidth}{!}{
                        \begin{tabular}{ccccc}  
                        \hline
                       \textbf{Scheme} & \textbf{L1 ↓} & \textbf{MSE ↓} & \textbf{LPIPS ↓}&\textbf{FID ↓}  \\ 
                        \hline
                        Sibling Attack~\cite{siblingattack} & 0.0411  & 0.0027  & 0.1218  &61.88  \\ 
                        AdvFaces~\cite{advfaces}       & \textbf{0.0199}  & \underline{0.0012}  & \underline{0.0429} &  69.79   \\ 
                        SemanticAdv~\cite{semanticadv}    & \underline{0.0284 } & \textbf{0.0016}  & \textbf{0.0395} & 114.61  \\ 
                        GMAA ~\cite{GMAA}          & 0.0984  & 0.0174  & 0.117 &  140.92  \\ 
                        \hline
                        ErasableMask (ours) & 0.0357  &  0.0028   &0.0645  &\textbf{19.08}  \\
                        Reconstructed (ours) & 0.0388& 0.0034 & 0.0981 &  \underline{55.89}\\
                        \hline
                        \end{tabular}
                        }
                        \label{quality}
                    \end{table}
        \subsection{Ablation Study}
        \subsubsection{Without Meta-auxiliary Attack}
        
                        In this section, we removed the meta-auxiliary attack while keeping other conditions unchanged. During the experiment, we observed that the adversarial loss $\mathcal{L}_{adv}$ rapidly declined, and concurrently, the quality of protected faces significantly dropped. The model quickly identified the local optimum for the three white-box models, thereby demonstrating strong adversarial capabilities on these models. However, this rapid decrease led to challenges for ErasableMask in finding more general and robust features of FR models, ultimately resulting in suboptimal black-box performance under ensemble attacks.
                        \begin{table}
                        \renewcommand{\arraystretch}{2.5}
                    \caption{Ablation study for Meta-auxiliary attack. ASR results for black-box offline FR models.}
                    \label{table_7}
                    \centering
                    \resizebox{0.5\textwidth}{!}{
                    \fontsize{15}{10}\selectfont
                    \begin{tabular}{cccccccc}
                    \hline

                    \multicolumn{2}{c}{\textbf{BlackBox}} & \multicolumn{1}{c}{\textbf{FaceNet}} & \multicolumn{1}{c}{\textbf{ArcFace}} & \multicolumn{1}{c}{\textbf{IRSE50}} & \multicolumn{1}{c}{\textbf{MobileFace}} & \multicolumn{1}{c}{\textbf{CosFace}} & \multicolumn{1}{c}{\textbf{IR152}} \\ \hline
                    \multicolumn{2}{c}{\multirow{1}{*}{w/o Meta-auxiliaryAttack}}  & \multicolumn{1}{c}{58.22} & \multicolumn{1}{c}{87.78} & \multicolumn{1}{c}{\textbf{95.56}} & \multicolumn{1}{c}{90.67} & \multicolumn{1}{c}{83.33} & 88.00 \\ 
                    \multicolumn{2}{c}{\multirow{1}{*}{Meta-auxiliary Attack (ours)}}  & \multicolumn{1}{c}{\textbf{64.00}} & \multicolumn{1}{c}{\textbf{90.44}} & \multicolumn{1}{c}{94.09} & \multicolumn{1}{c}{\textbf{92.22}} & \multicolumn{1}{c}{\textbf{93.31}} & \multicolumn{1}{c}{\textbf{92.00}} \\ 
                    \hline
                    \end{tabular}
                    }
                \end{table}
        \subsubsection{Without Information Injection}
                In this section, we conduct experiments on different information injection weights $\gamma$. The results are demonstrated in Fig.~\ref{loss} (a-b). As $\gamma$ increases, the adversarial performance decreases while the image quality and erasion enhance. Particularly, when $\gamma = 0.1$, the ASR results against Facenet and ArcFace drop to $56.67$ and $81.11$, respectively, with a decrease over $15$. The $\gamma$ demonstrates a trade-off between adversarial and erasion performance of ErasableMask, which can improve the ASR results via decreasing $\gamma$ and can improve the erasion performance through simply increasing $\gamma$. As shown in Fig.~\ref{loss} (a-b), ErasableMask without information injection demonstrates the best ASR results, while exhibiting the worst erasion ability and image quality. 
                Hence, ErasableMask is capable of utilizing various $\gamma$ to cater to diverse real-world circumstances.
        \subsubsection{ErasableMask with Different $\beta$}
\begin{figure*}[!t]
    \centering
    \footnotesize
    \begin{minipage}{0.3\textwidth}
        \centering
        \includegraphics[width=\linewidth]{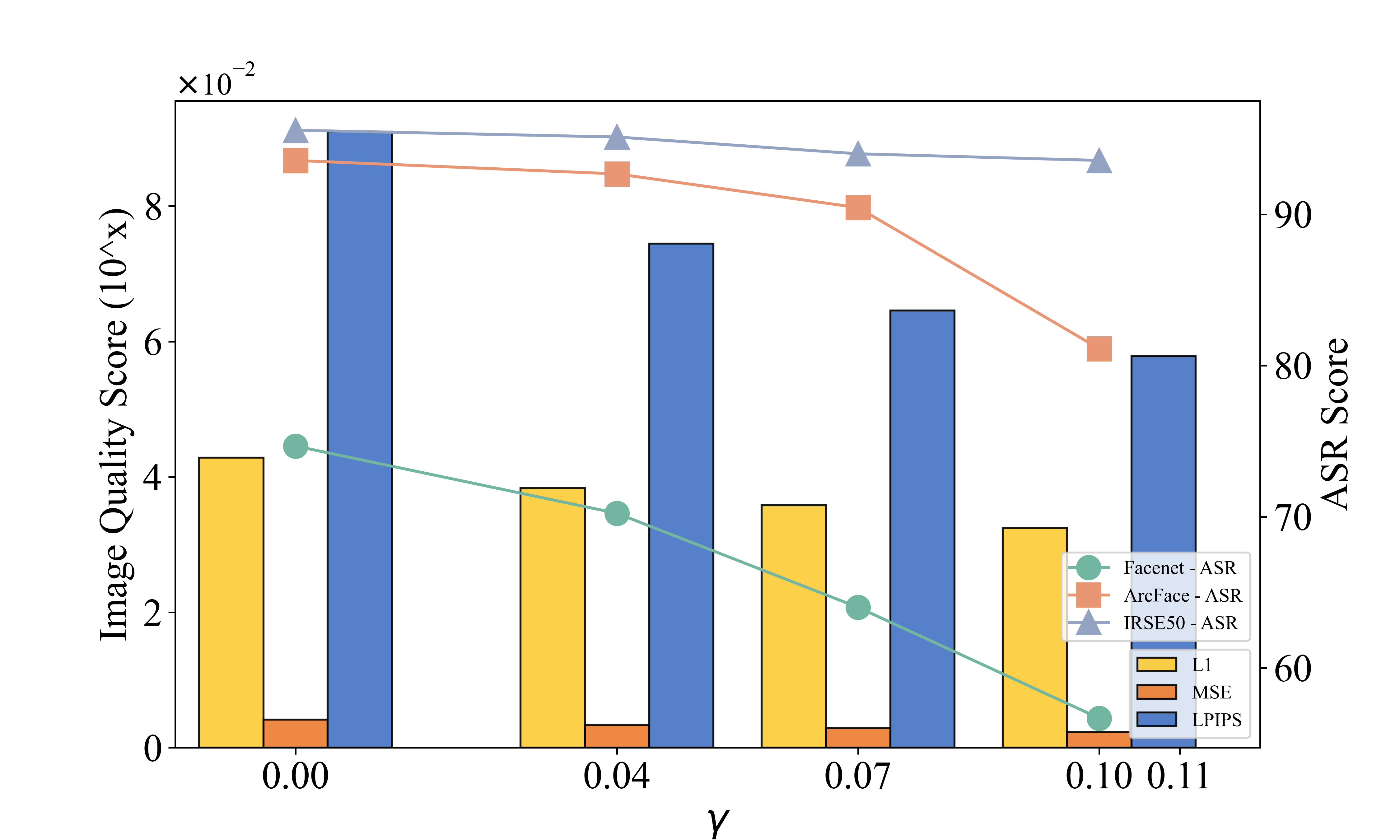}
        \vspace{1mm}
       a. The ASR results of ErasableMask with different $\gamma$
    \end{minipage}
    \begin{minipage}{0.3\textwidth}
        \centering
        \includegraphics[width=\linewidth]{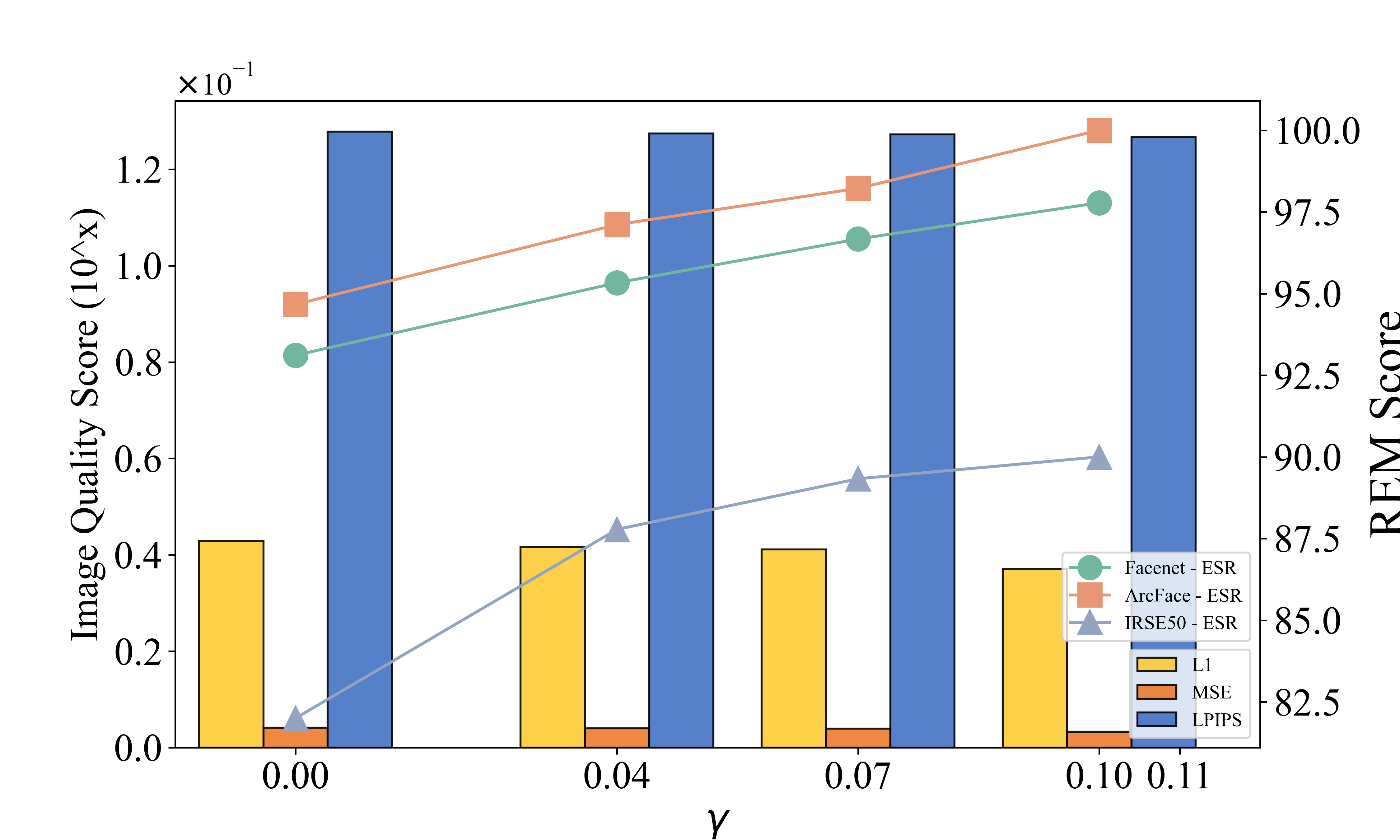}
        \vspace{1mm}
        b. The ESR results of ErasableMask with different $\gamma$
    \end{minipage}
    \begin{minipage}{0.3\textwidth}
        \centering
        \includegraphics[width=\linewidth]{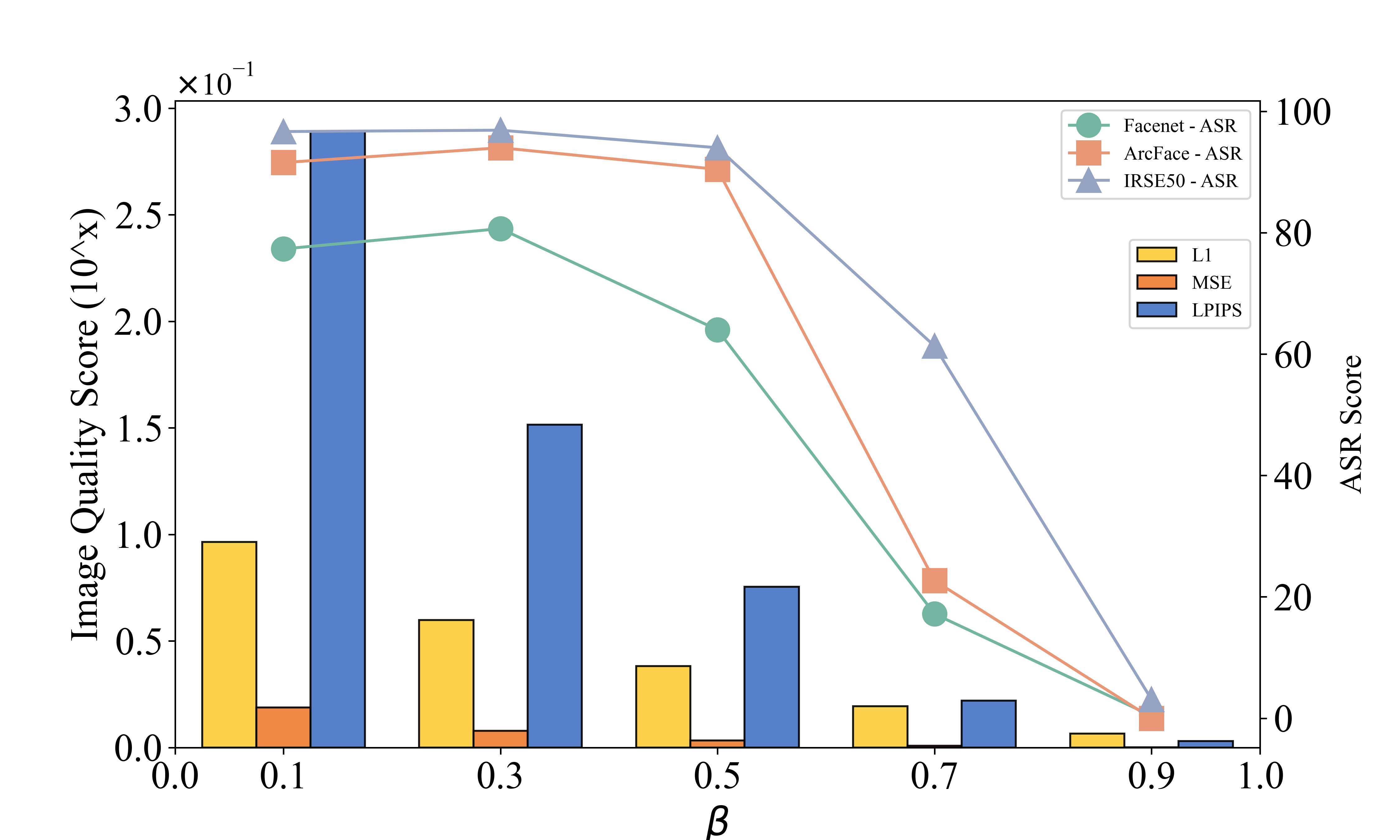}
        \vspace{1mm}
       c. ASR results and image quality with different $\beta$
    \end{minipage}
    \caption{\footnotesize ASR and ESR results for selection of different $\gamma$ and $\beta$.}
    \label{loss}
\end{figure*}

                In this section, we examine the selection of $\beta$. The experiment results are exhibited in Fig.~\ref{loss} (c). The ASR results tend to drop smoothly before $\beta$ decreases to $0.7$, where the image quality enhances sharply. Especially, when $\beta = 0.5$, the image quality has a significant advantage over those whose $\beta \leq 0.3$, while the ASR results show little decrease against ArcFace and IRSE50. Although there is a decrease in ASR results of Facenet, it is reasonable considering the enhancement of image quality. When $\beta \geq 0.7$, image quality quickly increases with a significant drop in ASR results. As a result, we set $\beta = 0.5$ in our whole experiment.
        

        \section{Conclusion and limitations}\label{sec.6}
        To address the limitations of existing facial privacy protection schemes, we propose ErasableMask, the first erasable adversarial example-based facial privacy protection scheme. ErasableMask introduces an erasable adversarial mask on facial images, preventing malicious hackers from conducting face verification while offers erasion mechanism for TAs. Specifically, we first adopt a novel meta-auxiliary attack to boost black-box transferability via fine-grained balance the optimization and contributions of different surrogate FR models. Then we utilize a self-erasable strategy based on clean-domain information injection to erase semantic perturbations in protected faces. We also employ a three-stage curriculum learning strategy to solve the optimization conflicts between adversarial and erasion performance. Through extensive experiments on CelebA-HQ and FFHQ datasets, ErasableMask demonstrates the state-of-the-art transferability, robustness, and erasion performance. However, current quality evaluation metrics for adversarial examples do not apply to scenarios based on semantic perturbations, making it impractical to simply assess naturalness via similarity between the adversarial example and the original face.
        
        \section{Acknowledgement}\label{sec.6}
        The authors would like to thank the editor and reviewers for their careful reading and valuable comments. This research was supported by National Natural Science Foundation of ChinaNSFC (No.62472325), and the Guangdong S\&T Program under Grant 2024B0101010002, and the National Natural Science Foundation of China (No.U2436208, No.62372129), and the Project of Guangdong Key Laboratory of Industrial Control System Security (2024B1212020010).
\bibliography{reference}

\begin{thebibliography}{10}
\providecommand{\url}[1]{#1}
\csname url@samestyle\endcsname
\providecommand{\newblock}{\relax}
\providecommand{\bibinfo}[2]{#2}
\providecommand{\BIBentrySTDinterwordspacing}{\spaceskip=0pt\relax}
\providecommand{\BIBentryALTinterwordstretchfactor}{4}
\providecommand{\BIBentryALTinterwordspacing}{\spaceskip=\fontdimen2\font plus
\BIBentryALTinterwordstretchfactor\fontdimen3\font minus \fontdimen4\font\relax}
\providecommand{\BIBforeignlanguage}[2]{{%
\expandafter\ifx\csname l@#1\endcsname\relax
\typeout{** WARNING: IEEEtran.bst: No hyphenation pattern has been}%
\typeout{** loaded for the language `#1'. Using the pattern for}%
\typeout{** the default language instead.}%
\else
\language=\csname l@#1\endcsname
\fi
#2}}
\providecommand{\BIBdecl}{\relax}
\BIBdecl

\bibitem{ding2015robust}
C.~Ding and D.~Tao, ``Robust face recognition via multimodal deep face representation,'' \emph{IEEE transactions on Multimedia}, vol.~17, no.~11, pp. 2049--2058, 2015.

\bibitem{zhong2021dynamic}
Y.~Zhong, W.~Deng, H.~Fang, J.~Hu, D.~Zhao, X.~Li, and D.~Wen, ``Dynamic training data dropout for robust deep face recognition,'' \emph{IEEE Transactions on Multimedia}, vol.~24, pp. 1186--1197, 2021.

\bibitem{chen2015face}
B.-C. Chen, C.-S. Chen, and W.~H. Hsu, ``Face recognition and retrieval using cross-age reference coding with cross-age celebrity dataset,'' \emph{IEEE Transactions on Multimedia}, vol.~17, no.~6, pp. 804--815, 2015.

\bibitem{choi2010collaborative}
J.~Y. Choi, W.~De~Neve, K.~N. Plataniotis, and Y.~M. Ro, ``Collaborative face recognition for improved face annotation in personal photo collections shared on online social networks,'' \emph{IEEE Transactions on Multimedia}, vol.~13, no.~1, pp. 14--28, 2010.

\bibitem{larson2018pixel}
M.~Larson, Z.~Liu, S.~Brugman, and Z.~Zhao, ``Pixel privacy. increasing image appeal while blocking automatic inference of sensitive scene information,'' 2018.

\bibitem{advfaces}
D.~Deb, J.~Zhang, and A.~K. Jain, ``Advfaces: Adversarial face synthesis,'' in \emph{2020 IEEE International Joint Conference on Biometrics (IJCB)}.\hskip 1em plus 0.5em minus 0.4em\relax IEEE, 2020, pp. 1--10.

\bibitem{komkov2021advhat}
S.~Komkov and A.~Petiushko, ``Advhat: Real-world adversarial attack on arcface face id system,'' in \emph{2020 25th international conference on pattern recognition (ICPR)}.\hskip 1em plus 0.5em minus 0.4em\relax IEEE, 2021, pp. 819--826.

\bibitem{dong2019efficient}
Y.~Dong, H.~Su, B.~Wu, Z.~Li, W.~Liu, T.~Zhang, and J.~Zhu, ``Efficient decision-based black-box adversarial attacks on face recognition,'' in \emph{proceedings of the IEEE/CVF conference on computer vision and pattern recognition}, 2019, pp. 7714--7722.

\bibitem{sharif2016accessorize}
M.~Sharif, S.~Bhagavatula, L.~Bauer, and M.~K. Reiter, ``Accessorize to a crime: Real and stealthy attacks on state-of-the-art face recognition,'' in \emph{Proceedings of the 2016 acm sigsac conference on computer and communications security}, 2016, pp. 1528--1540.

\bibitem{yang2021attacks}
L.~Yang, Q.~Song, and Y.~Wu, ``Attacks on state-of-the-art face recognition using attentional adversarial attack generative network,'' \emph{Multimedia tools and applications}, vol.~80, pp. 855--875, 2021.

\bibitem{semanticadv}
H.~Qiu, C.~Xiao, L.~Yang, X.~Yan, H.~Lee, and B.~Li, ``Semanticadv: Generating adversarial examples via attribute-conditioned image editing,'' in \emph{Computer Vision--ECCV 2020: 16th European Conference, Glasgow, UK, August 23--28, 2020, Proceedings, Part XIV 16}.\hskip 1em plus 0.5em minus 0.4em\relax Springer, 2020, pp. 19--37.

\bibitem{jia2022adv}
S.~Jia, B.~Yin, T.~Yao, S.~Ding, C.~Shen, X.~Yang, and C.~Ma, ``Adv-attribute: Inconspicuous and transferable adversarial attack on face recognition,'' \emph{Advances in Neural Information Processing Systems}, vol.~35, pp. 34\,136--34\,147, 2022.

\bibitem{hu2022protecting}
S.~Hu, X.~Liu, Y.~Zhang, M.~Li, L.~Y. Zhang, H.~Jin, and L.~Wu, ``Protecting facial privacy: Generating adversarial identity masks via style-robust makeup transfer,'' in \emph{Proceedings of the IEEE/CVF conference on computer vision and pattern recognition}, 2022, pp. 15\,014--15\,023.

\bibitem{shamshad2023clip2protect}
F.~Shamshad, M.~Naseer, and K.~Nandakumar, ``Clip2protect: Protecting facial privacy using text-guided makeup via adversarial latent search,'' in \emph{Proceedings of the IEEE/CVF Conference on Computer Vision and Pattern Recognition}, 2023, pp. 20\,595--20\,605.

\bibitem{liu2024adv}
D.~Liu, X.~Wang, C.~Peng, N.~Wang, R.~Hu, and X.~Gao, ``Adv-diffusion: imperceptible adversarial face identity attack via latent diffusion model,'' in \emph{Proceedings of the AAAI Conference on Artificial Intelligence}, vol.~38, no.~4, 2024, pp. 3585--3593.

\bibitem{hu2024towards}
C.~Hu, Y.~Li, Z.~Feng, and X.~Wu, ``Towards transferable attack via adversarial diffusion in face recognition,'' \emph{IEEE Transactions on Information Forensics and Security}, 2024.

\bibitem{yin2021adv}
B.~Yin, W.~Wang, T.~Yao, J.~Guo, Z.~Kong, S.~Ding, J.~Li, and C.~Liu, ``Adv-makeup: A new imperceptible and transferable attack on face recognition,'' in \emph{International Joint Conference on Artificial Intelligence}, 2021.

\bibitem{xiong2023black}
L.~Xiong, Y.~Wu, P.~Yu, and Y.~Zheng, ``A black-box reversible adversarial example for authorizable recognition to shared images,'' \emph{Pattern Recognition}, vol. 140, p. 109549, 2023.

\bibitem{liu2023unauthorized}
J.~Liu, W.~Zhang, K.~Fukuchi, Y.~Akimoto, and J.~Sakuma, ``Unauthorized ai cannot recognize me: Reversible adversarial example,'' \emph{Pattern Recognition}, vol. 134, p. 109048, 2023.

\bibitem{zhang2022self}
J.~Zhang, J.~Wang, H.~Wang, and X.~Luo, ``Self-recoverable adversarial examples: A new effective protection mechanism in social networks,'' \emph{IEEE Transactions on Circuits and Systems for Video Technology}, vol.~33, no.~2, pp. 562--574, 2022.

\bibitem{xie2024reversible}
Y.~Xie, Y.~Zhou, T.~Wang, W.~Wen, S.~Yi, and Y.~Zhang, ``Reversible gender privacy enhancement via adversarial perturbations,'' \emph{Neural Networks}, vol. 172, p. 106130, 2024.

\bibitem{yin2019reversible}
Z.~Yin, L.~Chen, W.~Lyu, and B.~Luo, ``Reversible attack based on adversarial perturbation and reversible data hiding in yuv colorspace,'' \emph{Pattern Recognition Letters}, vol. 166, pp. 1--7, 2023.

\bibitem{liu2019self}
S.~Liu, A.~Davison, and E.~Johns, ``Self-supervised generalisation with meta auxiliary learning,'' \emph{Advances in Neural Information Processing Systems}, vol.~32, 2019.

\bibitem{chen2024joint}
H.~Chen, X.~Wang, Y.~Zhou, Y.~Qin, C.~Guan, and W.~Zhu, ``Joint data-task generation for auxiliary learning,'' \emph{Advances in Neural Information Processing Systems}, vol.~36, 2024.

\bibitem{finn2017model}
C.~Finn, P.~Abbeel, and S.~Levine, ``Model-agnostic meta-learning for fast adaptation of deep networks,'' in \emph{International conference on machine learning}.\hskip 1em plus 0.5em minus 0.4em\relax PMLR, 2017, pp. 1126--1135.

\bibitem{shao2020regularized}
R.~Shao, X.~Lan, and P.~C. Yuen, ``Regularized fine-grained meta face anti-spoofing,'' in \emph{Proceedings of the AAAI conference on artificial intelligence}, vol.~34, no.~07, 2020, pp. 11\,974--11\,981.

\bibitem{jia2021mbrs}
Z.~Jia, H.~Fang, and W.~Zhang, ``Mbrs: Enhancing robustness of dnn-based watermarking by mini-batch of real and simulated jpeg compression,'' in \emph{Proceedings of the 29th ACM international conference on multimedia}, 2021, pp. 41--49.

\bibitem{curriculum}
X.~Wang, Y.~Chen, and W.~Zhu, ``A survey on curriculum learning,'' \emph{IEEE transactions on pattern analysis and machine intelligence}, vol.~44, no.~9, pp. 4555--4576, 2021.

\bibitem{fgsm}
I.~J. Goodfellow, ``Explaining and harnessing adversarial examples,'' \emph{arXiv preprint arXiv:1412.6572}, 2014.

\bibitem{pgd}
A.~Madry, A.~Makelov, L.~Schmidt, D.~Tsipras, and A.~Vladu, ``Towards deep learning models resistant to adversarial attacks,'' in \emph{6th International Conference on Learning Representations, {ICLR} 2018, Vancouver, BC, Canada, April 30 - May 3, 2018, Conference Track Proceedings}.\hskip 1em plus 0.5em minus 0.4em\relax OpenReview.net, 2018.

\bibitem{siblingattack}
Z.~Li, B.~Yin, T.~Yao, J.~Guo, S.~Ding, S.~Chen, and C.~Liu, ``Sibling-attack: Rethinking transferable adversarial attacks against face recognition,'' in \emph{Proceedings of the IEEE/CVF Conference on Computer Vision and Pattern Recognition}, 2023, pp. 24\,626--24\,637.

\bibitem{GMAA}
Q.~Li, Y.~Hu, Y.~Liu, D.~Zhang, X.~Jin, and Y.~Chen, ``Discrete point-wise attack is not enough: Generalized manifold adversarial attack for face recognition,'' in \emph{Proceedings of the IEEE/CVF Conference on Computer Vision and Pattern Recognition}, 2023, pp. 20\,575--20\,584.

\bibitem{dong2018boosting}
Y.~Dong, F.~Liao, T.~Pang, H.~Su, J.~Zhu, X.~Hu, and J.~Li, ``Boosting adversarial attacks with momentum,'' in \emph{Proceedings of the IEEE conference on computer vision and pattern recognition}, 2018, pp. 9185--9193.

\bibitem{xiao2021improving}
Z.~Xiao, X.~Gao, C.~Fu, Y.~Dong, W.~Gao, X.~Zhang, J.~Zhou, and J.~Zhu, ``Improving transferability of adversarial patches on face recognition with generative models,'' in \emph{Proceedings of the IEEE/CVF conference on computer vision and pattern recognition}, 2021, pp. 11\,845--11\,854.

\bibitem{yang2021towards}
X.~Yang, Y.~Dong, T.~Pang, H.~Su, J.~Zhu, Y.~Chen, and H.~Xue, ``Towards face encryption by generating adversarial identity masks,'' in \emph{Proceedings of the IEEE/CVF International Conference on Computer Vision}, 2021, pp. 3897--3907.

\bibitem{zhong2020towards}
Y.~Zhong and W.~Deng, ``Towards transferable adversarial attack against deep face recognition,'' \emph{IEEE Transactions on Information Forensics and Security}, vol.~16, pp. 1452--1466, 2020.

\bibitem{carlini2017towards}
N.~Carlini and D.~Wagner, ``Towards evaluating the robustness of neural networks,'' in \emph{2017 ieee symposium on security and privacy (sp)}.\hskip 1em plus 0.5em minus 0.4em\relax Ieee, 2017, pp. 39--57.

\bibitem{wei2022simultaneously}
X.~Wei, Y.~Guo, J.~Yu, and B.~Zhang, ``Simultaneously optimizing perturbations and positions for black-box adversarial patch attacks,'' \emph{IEEE transactions on pattern analysis and machine intelligence}, vol.~45, no.~7, pp. 9041--9054, 2022.

\bibitem{li2024transferable}
M.~Li, J.~Wang, H.~Zhang, Z.~Zhou, S.~Hu, and X.~Pei, ``Transferable adversarial facial images for privacy protection,'' \emph{arXiv preprint arXiv:2408.01428}, 2024.

\bibitem{he2019attgan}
Z.~He, W.~Zuo, M.~Kan, S.~Shan, and X.~Chen, ``Attgan: Facial attribute editing by only changing what you want,'' \emph{IEEE transactions on image processing}, vol.~28, no.~11, pp. 5464--5478, 2019.

\bibitem{karras2017progressive}
T.~Karras, ``Progressive growing of gans for improved quality, stability, and variation,'' \emph{arXiv preprint arXiv:1710.10196}, 2017.

\bibitem{karras2019style}
T.~Karras, S.~Laine, and T.~Aila, ``A style-based generator architecture for generative adversarial networks,'' in \emph{Proceedings of the IEEE/CVF conference on computer vision and pattern recognition}, 2019, pp. 4401--4410.

\bibitem{zhang2016joint}
K.~Zhang, Z.~Zhang, Z.~Li, and Y.~Qiao, ``Joint face detection and alignment using multitask cascaded convolutional networks,'' \emph{IEEE signal processing letters}, vol.~23, no.~10, pp. 1499--1503, 2016.

\bibitem{he2016deep}
K.~He, X.~Zhang, S.~Ren, and J.~Sun, ``Deep residual learning for image recognition,'' in \emph{Proceedings of the IEEE conference on computer vision and pattern recognition}, 2016, pp. 770--778.

\bibitem{hu2018squeeze}
J.~Hu, L.~Shen, and G.~Sun, ``Squeeze-and-excitation networks,'' in \emph{Proceedings of the IEEE conference on computer vision and pattern recognition}, 2018, pp. 7132--7141.

\bibitem{schroff2015facenet}
F.~Schroff, D.~Kalenichenko, and J.~Philbin, ``Facenet: A unified embedding for face recognition and clustering,'' in \emph{Proceedings of the IEEE conference on computer vision and pattern recognition}, 2015, pp. 815--823.

\bibitem{deng2019arcface}
J.~Deng, J.~Guo, N.~Xue, and S.~Zafeiriou, ``Arcface: Additive angular margin loss for deep face recognition,'' in \emph{Proceedings of the IEEE/CVF conference on computer vision and pattern recognition}, 2019, pp. 4690--4699.

\bibitem{duta2021improved}
I.~C. Duta, L.~Liu, F.~Zhu, and L.~Shao, ``Improved residual networks for image and video recognition,'' in \emph{2020 25th International Conference on Pattern Recognition (ICPR)}.\hskip 1em plus 0.5em minus 0.4em\relax IEEE, 2021, pp. 9415--9422.

\bibitem{wang2018cosface}
H.~Wang, Y.~Wang, Z.~Zhou, X.~Ji, D.~Gong, J.~Zhou, Z.~Li, and W.~Liu, ``Cosface: Large margin cosine loss for deep face recognition,'' in \emph{Proceedings of the IEEE conference on computer vision and pattern recognition}, 2018, pp. 5265--5274.

\bibitem{MEGVII}
MEGVII, ``In https://www.faceplusplus.com.cn/,'' 2021.

\bibitem{tencentapi}
T.~Cloud, \url{https://cloud.tencent.com/document/product/867}.

\bibitem{Aliyun}
Aliyun, ``https://cn.aliyun.com/,'' 2019.

\bibitem{madry2017towards}
A.~Madry, ``Towards deep learning models resistant to adversarial attacks,'' \emph{arXiv preprint arXiv:1706.06083}, 2017.

\bibitem{heusel2018ganstrainedtimescaleupdate}
M.~Heusel, H.~Ramsauer, T.~Unterthiner, B.~Nessler, and S.~Hochreiter, ``Gans trained by a two time-scale update rule converge to a local nash equilibrium,'' \emph{Advances in neural information processing systems}, vol.~30, 2017.

\bibitem{zhang2018unreasonableeffectivenessdeepfeatures}
R.~Zhang, P.~Isola, A.~A. Efros, E.~Shechtman, and O.~Wang, ``The unreasonable effectiveness of deep features as a perceptual metric,'' in \emph{CVPR}, 2018.

\end{thebibliography}
	\ifCLASSOPTIONcaptionsoff
	\newpage
	\fi
	\bibliographystyle{IEEEtran}
\end{document}